\documentclass{article}
     \PassOptionsToPackage{round,compress}{natbib}
\usepackage[final]{neurips_2022}

\usepackage{amsmath,amsfonts,bm}
\usepackage{mathtools}

\newcommand{\ba}[1]{\begin{align}#1\end{align}}

\newcommand{\given}{\,|\,}

\def\eqref#1{equation~\ref{#1}}

\def\1{\bm{1}}

\def\rvtheta{{\mathbf{\theta}}}

\def\vmu{{\bm{\mu}}}

\def\vx{{\bm{x}}}
\def\vy{{\bm{y}}}
\def\vz{{\bm{z}}}

\def\mI{{\bm{I}}}

\DeclareMathAlphabet{\mathsfit}{\encodingdefault}{\sfdefault}{m}{sl}
\SetMathAlphabet{\mathsfit}{bold}{\encodingdefault}{\sfdefault}{bx}{n}

\def\gL{{\mathcal{L}}}

\newcommand{\KL}{D_{\mathrm{KL}}}

\DeclareMathOperator*{\argmax}{arg\,max}

\DeclarePairedDelimiterX{\infdivx}[2]{(}{)}{%
  #1\;\delimsize\|\;#2%
}
\newcommand{\kldiv}{D_{\mathrm{KL}}\infdivx}

 \newcommand{\Sigmamat}[0]{\ensuremath{\boldsymbol{\Sigma}} }
 \newcommand{\epsilonv}[0]{\ensuremath{\boldsymbol{\epsilon}} }
\newcommand*\circled[1]{\tikz[baseline=(char.base)]{
            \node[shape=circle,draw,inner sep=2pt] (char) {#1};}}

\usepackage[utf8]{inputenc} 
\usepackage[T1]{fontenc}    
\usepackage{hyperref}       
\usepackage{url}            
\usepackage{booktabs}       
\usepackage{amsfonts}       
\usepackage{nicefrac}       
\usepackage{microtype}      
\usepackage{xcolor}         
\usepackage{bbold}
\usepackage{algorithm}
\usepackage{algorithmic}
\usepackage{makecell}
\usepackage{tikz}
\usepackage{stackengine}
\usepackage{multibib}
\usepackage{wrapfig}
\newcites{apndx}{References}
\usepackage{multirow}
\usepackage{tablefootnote}

\newcommand*\samethanks[1][\value{footnote}]{\footnotemark[#1]}

\title{
CARD: Classification and Regression Diffusion Models
}

\author{
  Xizewen Han\thanks{Equal contribution.}~~~~ Huangjie Zheng\samethanks \\
  Department of Statistics and Data Sciences\\
  The University of Texas at Austin\\
  Austin, TX 78712 \\
  \texttt{\{xizewen.han,\,huangjie.zheng\}@utexas.edu}\!
\And
Mingyuan Zhou \\
McCombs School of Business\\
  The University of Texas at Austin\\
  Austin, TX 78712 \\
\texttt{mingyuan.zhou@mccombs.utexas.edu}
}

\begin{document}

\maketitle

\begin{abstract}
Learning the distribution of a continuous or categorical response variable $\vy$ given its covariates $\vx$ is a fundamental problem in statistics and machine learning. Deep neural network-based supervised learning algorithms have made great progress in predicting the mean of $\vy$ given $\vx$, but they are often criticized for their ability to accurately capture the uncertainty of their predictions. In this paper, we introduce classification and regression diffusion (CARD) models, which combine a denoising diffusion-based conditional generative model and a pre-trained conditional mean estimator, to accurately predict the distribution of $\vy$ given $\vx$. We demonstrate the outstanding ability of CARD in conditional distribution prediction with both toy examples and real-world datasets, the experimental results on which show that CARD in general outperforms state-of-the-art methods, including Bayesian neural network-based ones that are designed for uncertainty estimation, especially when the conditional distribution of $\vy$ given $\vx$ is multi-modal. In addition, we utilize the stochastic nature of the generative model outputs to obtain a finer granularity in model confidence assessment at the instance level for classification tasks. Our implementation is publicly available at \href{https://github.com/XzwHan/CARD}{https://github.com/XzwHan/CARD}.
\end{abstract}

\section{Introduction}\label{sec:intro}
A fundamental problem in statistics and machine learning is to predict the response variable $\vy$ given a set of covariates $\vx$. Generally speaking, $\vy$ is a continuous variable for regression analysis and a  categorical variable for classification. Denote $f(\vx)\in \mathbb{R}^C$ as a deterministic function that transforms $\vx$ into a $C$ dimensional output. Denote $f_c(\vx)$ as the $c$-th dimension of $f(\vx)$. Existing methods typically assume an additive noise model: for regression analysis with $\vy\in\mathbb{R}^C$, one often assumes $\vy=f(\vx)+\epsilonv,~\epsilonv\sim \mathcal{N}(0,\Sigmamat)$, while for classification with $y\in\{1,\ldots,C\}$, one often assumes $y=\argmax\big(f_1(\vx)+\epsilonv_1,\ldots,f_C(\vx)+\epsilonv_C\big)$, where  $\epsilonv_c\stackrel{iid}\sim \mbox{EV}_1(0,1)$, a standard type-1 extreme value distribution. Thus we have the expected value of $\vy$ given $\vx$ as $\mathop{\mathbb{E}}[\vy\given \vx] = f(\vx)$ in regression and $P(y=c\given \vx)=\mathbb{E}[y=c\given \vx] = \mbox{softmax}_c\big(f(\vx)\big)=\frac{\exp\left({f_c(\vx)}\right)}{\sum_{c'=1}^C \exp\left({f_{c'}(\vx)}\right) }$ in classification.

These additive-noise models are primarily focusing on accurately estimating the conditional mean $\mathbb{E}[\vy\given \vx]$, while paying less attention to whether the noise distribution can accurately capture the uncertainty of $\vy$ given $\vx$. For this reason, they may not work well if the distribution of $\vy$ given $\vx$ clearly deviates from the additive-noise assumption. For example, if $p(\vy\given \vx)$ is multi-modal, which commonly happens when there are missing categorical covariates in $\vx$, then $\mathbb{E}[\vy\given \vx]$ may not be close to any possible true values of $\vy$ given that specific $\vx$. More specifically, consider a person whose weight, height, blood pressure, and age are known but gender is unknown, then the testosterone or estrogen level of this person is likely to follow a bi-modal distribution and the chance of developing breast cancer is also likely to follow a bi-modal distribution. Therefore, these widely used additive-noise models, which use a deterministic function $f(\vx)$ to characterize the conditional mean of $\vy$, are inherently restrictive in their ability for uncertainty estimation.

In this paper, our goal is to accurately recover the full distribution of $\vy$ conditioning on $\vx$ given a set of $N$ training data points, denoted as $\mathcal{D}=\{(\vx_i,\vy_i)\}_{1,N}$. To realize this goal, we consider the diffusion-based (a.k.a. score-based) generative models \citep{diffusion,scorematching,ddpm,improvedscore,song2021scorebased} and inject covariate-dependence into both the forward and reverse diffusion chains. Our method can model the conditional distribution of both continuous and categorical $\vy$ variables, and the algorithms developed under this method will be collectively referred to as \textbf{C}lassification \textbf{A}nd \textbf{R}egression \textbf{D}iffusion (CARD) models.

Diffusion-based generative models have received significant recent attention due to not only their ability to generate high-dimensional data, such as high-resolution photo-realistic images, but also their training stability.
They can be understood from the perspective of score matching \citep{hyvarinen2005estimation,vincent2011connection} and Langevin dynamics \citep{neal2011mcmc,welling2011bayesian}, as pioneered by \citet{scorematching}. They can also be understood from the perspective of diffusion probabilistic models~\citep{diffusion,ddpm}, which first define a forward diffusion to transform the data into noise and then a reverse diffusion to regenerate the data from~noise.

These previous methods mainly focus on unconditional generative modeling. While there exist guided-diffusion models \citep{scorematching,song2021scorebased,Dhariwal2021DiffusionMB,nichol2021glide,ramesh2022hierarchical} that target on generating high-resolution photo-realistic images that match the semantic meanings or content of the label, text, or corrupted-images, we focus on studying diffusion-based conditional generative modeling at a more fundamental level. In particular, our goal is to thoroughly investigate whether CARD can help accurately recover $p(\vy\given \vx,\mathcal D)$, the predictive distribution of $\vy$ given $\vx$ after observing data $\mathcal D =\{(\vx_i,\vy_i)\}_{i=1,N}$. In other words, our focus is on regression analysis of continuous or categorical response variables given their corresponding~covariates.

We summarize our main contributions as follows: 1) We show CARD, which injects covariate-dependence and a pre-trained conditional mean estimator into both the forward and reverse diffusion chains to construct a denoising diffusion probabilistic model, provides an accurate estimation of $p(\vy\given \vx,\mathcal D)$. 2) We provide a new metric to better evaluate how well a regression model captures the full distribution $p(\vy\given \vx,\mathcal D)$. 3) Experiments on standard benchmarks for regression analysis show that CARD achieves state-of-the-art results, using both existing metrics and the new one. 4) For classification tasks, we push the assessment of model confidence in its predictions towards the level of individual~instances, a finer~granularity than the previous methods.

\section{Methods and Algorithms for CARD}
Given the ground-truth response variable $\vy_0$ and its covariates $\vx$, and assuming a sequence of intermediate predictions $\vy_{1:T}$ made by the diffusion model, the goal of supervised learning is to learn a model such that the log-likelihood is maximized by optimizing the following ELBO: 
\ba{
\log p_{\rvtheta}(\vy_0 \given \vx)  = 
 \log \int p_{\rvtheta}( \vy_{0:T}\given \vx) d\vy_{1;T} \geq \mathbb{E}_{q(\vy_{1:T}\given\vy_0, \vx)}\left[\log \frac{p_{\rvtheta}(\vy_{0:T}\given \vx)}{q(\vy_{1:T}\given\vy_0, \vx)} \right], 
}
where $q(\vy_{1:T}\given \vy_0, \vx)$ is called the forward process or diffusion process in the concept of diffusion models~\citep{diffusion,ddpm}. Denote $\KL(q\text{ }||\text{ }p)$ as the Kullback--Leibler (KL) divergence from distribution $p$ to distribution $q$. The above objective can be rewritten as
\ba{
    \gL_{\text{ELBO}}(\vy_0, \vx) &\coloneqq \gL_0(\vy_0, \vx) + \sum_{t=2}^T\gL_{t-1}(\vy_0, \vx) + \gL_T(\vy_0, \vx) \label{eq:regression_loss}, \\
                     \gL_0(\vy_0, \vx) &\coloneqq \mathbb{E}_{q} \left[-\log p_{\theta}(\vy_0 \given  \vy_1, \vx)\right] \label{eq:regression_loss0}, \\
                     \gL_{t-1}(\vy_0, \vx)&\coloneqq 
                     \mathbb{E}_{q} \left[ \kldiv[\big]{q(\vy_{t-1}\given \vy_t,\vy_0, \vx)}{p_{\theta}(\vy_{t-1}\given \vy_t, \vx)} \right] \label{eq:regression_losst}, \\
                     \gL_T(\vy_0, \vx)&\coloneqq  \mathbb{E}_{q} \left[ \kldiv[\big]{q(\vy_T \given  \vy_0, \vx)}{p(\vy_T \given \vx)} \right] \label{eq:regression_lossT}.
}
Here we follow the convention to assume $\gL_T$ does not depend on any parameter and it will be close to zero by carefully diffusing the observed response variable $\vy_0$ towards a pre-assumed distribution $p(\vy_T\given\vx)$. The remaining terms will make the model $p_{\theta}(\vy_{t-1}\given \vy_t, \vx)$ approximate the corresponding tractable ground-truth denoising transition step $q(\vy_{t-1}\given \vy_t,\vy_0, \vx)$ for all timesteps. Different from the vanilla diffusion models, we assume the endpoint of our diffusion process to be
\ba{
    p(\vy_T\given\vx)=\mathcal{N}(f_{\phi}(\vx), \mI) \label{eq:prior},
}
where $f_{\phi}(\vx)$ is the prior knowledge of the relation between $\vx$ and $\vy_0$, \textit{e.g.}, a network pre-trained with $\mathcal D$ to approximate $\mathbb E[\vy\given \vx]$, or $\mathbf{0}$ if we assume the relation is unknown.
With a diffusion schedule $\{\beta_t\}_{t=1:T} \in (0,1)^T$, we specify the forward process conditional distributions in a similar fashion as \citet{pandey2022diffusevae}, but for all timesteps including $t=1$: 
\ba{
    q\big(\vy_t\given\vy_{t-1}, f_{\phi}(\vx)\big)=\mathcal{N}\big(\vy_t; \sqrt{1-\beta_t}\vy_{t-1} + (1-\sqrt{1-\beta_t})f_{\phi}(\vx), \beta_t\mI\big), \label{eq:form_1_forward_one_step}
}
which admits a closed-form sampling distribution with an arbitrary timestep $t$:
\ba{
    q\big(\vy_t\given\vy_0, f_{\phi}(\vx)\big)=\mathcal{N}\big(\vy_t; \sqrt{\bar{\alpha}_t}\vy_0 + (1-\sqrt{\bar{\alpha}_t})f_{\phi}(\vx), (1-\bar{\alpha}_t)\mI\big), \label{eq:form_1_forward_t_steps}
}
where $\alpha_t := 1-\beta_t$ and $\bar{\alpha}_t:=\prod_t\alpha_t$. Note that the mean term in Eq.~(\ref{eq:form_1_forward_one_step}) can be viewed as an interpolation between true data $\vy_0$ and the predicted conditional expectation $f_{\phi}(\vx)$, which gradually changes from the former to the latter throughout the forward process.

Such formulation corresponds to a tractable forward process posterior: 
\ba{
    q(\vy_{t-1}\given\vy_t, \vy_0, \vx)=q\big(\vy_{t-1}\given\vy_t, \vy_0, f_{\phi}(\vx)\big)=\mathcal{N}\Big(\vy_{t-1}; \tilde{\vmu}\big(\vy_t, \vy_0, f_{\phi}(\vx)\big), \tilde{\beta_t}\mI\Big), \label{eq:form_1_posterior}
}
where
\ba{
     \tilde{\vmu}&\coloneqq\underbrace{\frac{\beta_t\sqrt{\bar{\alpha}_{t-1}}}{1-\bar{\alpha}_t}}_{\gamma_0}\vy_0+\underbrace{\frac{(1-\bar{\alpha}_{t-1})\sqrt{\alpha_t}}{1-\bar{\alpha}_t}}_{\gamma_1}\vy_t+\underbrace{\Bigg(1+\frac{(\sqrt{\bar{\alpha}_t}-1)(\sqrt{\alpha_t}+\sqrt{\bar{\alpha}_{t-1}})}{1-\bar{\alpha}_t}\Bigg)}_{\gamma_2}f_{\phi}(\vx), \notag \\
    \tilde{\beta_t}&\coloneqq\frac{1-\bar{\alpha}_{t-1}}{1-\bar{\alpha}_t}\beta_t. %
    \notag
}
We provide the derivation in Appendix \ref{ssec:fwd_process_post_derivation}. The labels under the terms are used in Algorithm~\ref{alg:reg_form_1_inf}.

\subsection{CARD for Regression}\label{ssec:card_reg}
For regression problems, the goal of the reverse diffusion process is to gradually recover the distribution of the noise term, the aleatoric or local uncertainty inherent in the observations  \citep{alex2017uncertainty,wang2020thompson}, enabling us to generate samples that match the true conditional $p(\vy\given\vx)$.

Following the reparameterization introduced by denoising diffusion probabilistic models (DDPM) \citep{ddpm}, we construct $\epsilonv_{\rvtheta}\big(\vx, \vy_t, f_{\phi}(\vx), t\big)$, which is a function approximator parameterized by a deep neural network that predicts the forward diffusion noise $\epsilonv$ sampled for~$\vy_t$. The training and inference procedure can be carried out in a standard DDPM manner.

\begin{algorithm}[ht]
\begin{algorithmic}[1]
\STATE Pre-train $f_{\phi}(\vx)$ that predicts $\mathbb{E}(\vy\given\vx)$ with MSE
\REPEAT
\STATE Draw $y_0\sim q(\vy_0\given\vx)$
\STATE Draw $t\sim \text{Uniform}(\{1\dots T\})$
\STATE Draw $\epsilonv\sim\mathcal{N}(\mathbf{0}, \mI)$
\STATE Compute noise estimation loss $$\mathcal{L}_{\epsilonv} = \big|\big|\epsilonv - \epsilonv_{\rvtheta}\big(\vx,\sqrt{\bar{\alpha}_t}\vy_0+\sqrt{1-\bar{\alpha}_t}\epsilonv+(1-\sqrt{\bar{\alpha}_t}) f_{\phi}(\vx), f_{\phi}(\vx), t\big)\big|\big|^2$$
\STATE Take numerical optimization step on: $$\nabla_{\theta} \mathcal{L}_{\epsilonv} $$
\UNTIL Convergence
\end{algorithmic}
\caption{Training (Regression)}
\label{alg:reg_form_1_train}
\end{algorithm}

\begin{algorithm}[ht]
\begin{algorithmic}[1]
\STATE $\vy_T\sim\mathcal{N}(f_{\phi}(\vx), \mI)$
\FOR{$t=T$ to $1$}
\STATE Draw $\vz\sim\mathcal{N}(\bm{0},\mI) $if $t>1$
\STATE Calculate reparameterized $\hat{\vy}_0=\frac{1}{\sqrt{\bar{\alpha}_t}}\Big(\vy_t-(1-\sqrt{\bar{\alpha}_t})f_{\phi}(\vx)-\sqrt{1-\bar{\alpha}_t}\epsilonv_{\rvtheta}\big(\vx, \vy_t, f_{\phi}(\vx), t\big)\Big)$
\STATE Let $\vy_{t-1}=\gamma_0\hat{\vy}_0+\gamma_1\vy_t+\gamma_2f_{\phi}(\vx)+\sqrt{\tilde{\beta}_t}\vz$ if $t>1$, else set $\vy_{t-1}=\hat{\vy}_0$
\ENDFOR
\STATE \textbf{return} $\vy_0$
\end{algorithmic}
\caption{Inference (Regression)}
\label{alg:reg_form_1_inf}
\end{algorithm}

\subsection{CARD for Classification}\label{ssec:card_cls}
We formulate the classification tasks in a similar fashion as in Section~\ref{ssec:card_reg}, where we:
\begin{enumerate}
\item Replace the continuous response variable with a one-hot encoded label vector for $\vy_0$;
\item Replace the mean estimator with a pre-trained classifier that outputs softmax probabilities of the class labels for $f_{\phi}(\vx)$.
\end{enumerate}

This construction no longer assumes $\vy_0$ to be drawn from a categorical distribution, but instead treats each one-hot label as a class prototype, \textit{i.e.}, we assume a continuous data and state space, which enables us to keep the Gaussian diffusion model framework. The sampling procedure would output reconstructed $\vy_0$ in the range of real numbers for each dimension, instead of a vector in the probability simplex. Denoting $C$ as the number of classes and $\bm{1}_C$ as a $C$-dimensional vector of $1$s, we convert such output to a probability vector in a softmax form of a temperature-weighted Brier score \citep{brierscore}, which computes the squared error between the prediction and $\bm{1}_C$.
Mathematically, the probability of predicting the $k^{th}$ class and the final point prediction $\hat y$  can be expressed~as
\ba{
    \text{Pr}(y = k)=\frac{\exp(-(\vy_{0}-\bm{1}_C)^2_{k} / \tau \big)}{\sum_{i=1}^C \exp(-(\vy_{0}-\bm{1}_C)^2_{i} / \tau \big)}; ~ \hat y=\argmax_k \big(-(\vy_0-\bm{1}_C)^2_k\big). \label{eq:cls_output_conversion}
}
where $\tau > 0$ is the temperature parameter, and $(\vy_{0}-\bm{1}_C)^2_{k}$ indicates the $k^{th}$ dimension of the vector of element-wise square error between $\vy_{0}$ and $\bm{1}_C$, \textit{i.e.}, $(\vy_{0}-\bm{1}_C)^2_{k}= \| \vy_{0_{k}} - 1\|^2$.
Intuitively, this construction would assign the class whose raw output in the sampled $\vy_0$ is closest to the true class, encoded by the value of $1$ in the one-hot label, with the highest probability.

Conditional on the same covariates $\vx$, the stochasticity of the generative model would give us a different class prototype reconstruction after each reverse process sampling, which enables us to construct predicted probability intervals for all class labels. Such stochastic reconstruction is in a similar fashion as DALL-E $2$ %
\citep{ramesh2022hierarchical}
that applies a diffusion prior to reconstruct the image embedding by conditioning on the text embedding during the reverse diffusion process, which is a key step in the diversity of generated images.

\section{Related Work}\label{sec:relatedwork}
Under the supervised learning settings, to model the conditional distribution $p(\vy \given \vx)$ besides just the conditional mean $\mathop{\mathbb{E}}[\vy\given \vx]$ through deep neural networks, existing works have been focusing on quantifying predictive uncertainty, and several lines of work have been proposed. Bayesian neural networks (BNNs) model such uncertainty by assuming distributions over network parameters, capturing the plausibility of the model given the data \citep{bbb, pbp, mcdropout, vdropout, newelbo}. \citet{alex2017uncertainty} also model the uncertainties in the model outputs besides model parameters, by including the additive noise term as part of the neural network output. Meanwhile, ensemble-based methods \citep{deepensembles, deepensemblesrecent} have been proposed to model predictive uncertainty by combining multiple neural networks with stochastic outputs. Furthermore, the Neural Processes Family \citep{np, cnp, anp, convcnp} has introduced a series of models that capture predictive uncertainty in an out-of-distribution fashion, particularly designed for few-shot learning settings.

These above mentioned models have all assumed a parametric form in $p(\vy \given \vx)$, namely Gaussian distribution, or a mixture of Gaussians, and optimize the network parameters based on a Gaussian negative log-likelihood objective function. Deep generative models, on the other hand, have been known for modeling implicit distributions without parametric distributional assumptions, but very few works have been proposed to utilize such feature to tackle regression tasks. GAN-based models are introduced by \citet{jointmatching} and \citet{wganjointmatching} for conditional density estimation and predictive uncertainty quantification. For classification tasks, on the other hand, generative classifiers \citep{genclfearlywork, genclf1, genclf2, genclf3} is a class of models that also perform classification with generative models; among them, \citet{genclf} propose score-based generative classifiers to tackle classification tasks with score-based generative models \citep{song2021maximum, song2021scorebased}. They model $p(\vx \given \vy)$ and predict the label with the largest conditional likelihood of $\vx$, while CARD models $p(\vy \given \vx)$ instead.

In recent years, the class of diffusion-based (or score-based) deep generative models has demonstrated its outstanding performance in modeling high-dimensional multi-modal distributions \citep{ddpm, ddim, kawar2022denoising, xiao2021tackling, Dhariwal2021DiffusionMB, scorematching, improvedscore}, with most work focusing on Gaussian diffusion processes operating in continuous state spaces. \citet{multimomialdiffusion} introduce extensions of diffusion models for categorical data, and \citet{austin2021structured} have proposed diffusion models for discrete data as a generalization of the multinomial diffusion models, which could provide an alternative way of performing classification with diffusion-based models.

\section{Experiments}\label{sec:card_experiments}
For the hyperparameters of CARD in both regression and classification tasks, we set the number of timesteps as $T=1000$, a linear noise schedule with $\beta_1=10^{-4}$ and $\beta_T=0.02$, same as \citet{ddpm}. We provide a more detailed walk-through of the experimental setup, including training and network architecture, in Appendix \ref{ssec:general_experiment_details}.

\subsection{Regression}
Putting aside its statistical interpretation, the word \textit{regress} indicates a direction opposite to \textit{progress}, suggesting a less developed state. Such semantics in fact translates well into the statistical domain, in the sense that traditional regression analysis methods %
often only focus on estimating  $\mathbb{E}(\vy\given\vx)$, while leaving out all remaining details about $p(\vy\given\vx)$. In recent years, Bayesian neural networks (BNNs) have emerged as a class of models that aims at estimating the uncertainty \citep{pbp, mcdropout, deepensembles, newelbo}, providing a more complete picture of $p(\vy\given\vx)$. The metric that they use to quantify uncertainty estimation, negative log-likelihood (NLL), is computed with a Gaussian density, implying their assumption such that the conditional distributions $p(\vy\given\vx = x)$ for all $x$ are Gaussian. However, this assumption is very difficult to verify for real-world datasets: the covariates can be arbitrarily high-dimensional, making the feature space increasingly sparse with respect to the number of collected~observations.

To accommodate the need for uncertainty estimation without imposing such restriction for the parametric form of $p(\vy\given\vx)$, we apply the following two metrics, both of which are designed to empirically evaluate the level of similarity between the learned and the true conditional distributions:
\begin{enumerate}
    \item Prediction Interval Coverage Probability (PICP);
    \item Quantile Interval Coverage Error (QICE).
\end{enumerate}
PICP has been described in \citet{bnnquality}, whereas QICE is a new metric proposed by us. We describe both of them in what follows.

\subsubsection{PICP and QICE}
The PICP is computed as
\ba{
    \text{PICP}\coloneqq\frac{1}{N}\sum_{n=1}^N\mathbb{1}_{y_n\geq\hat{y}_n^{\text{low}}}\cdot\mathbb{1}_{y_n\leq\hat{y}_n^{\text{high}}}, \label{eq:PICP}
}
where $\hat{y}_n^{\text{low}}$ and $\hat{y}_n^{\text{high}}$ represent the low and high percentiles, respectively, of our choice for the predicted $\vy$ outputs given the same $\vx$ input. This metric measures the proportion of true observations that fall in the percentile range of the generated $\vy$ samples given each $\vx$ input. Intuitively, when the learned distribution represents the true distribution well, this measurement should be close to the difference between the selected low and high percentiles. In this paper, we choose the $2.5^{th}$ and $97.5^{th}$ percentile, thus an ideal PICP value for the learned model should be~$95\%$.

Meanwhile, there is a caveat for this metric: for example, imagine a situation where the $2.5^{th}$ to $97.5^{th}$ percentile of the learned distribution happens to cover the data between the $1^{st}$ and $96^{th}$ percentiles from the true distribution. Given enough samples, we shall still obtain a PICP value close to $95\%$, but clearly there is a mismatch between the learned distribution and the true~one.

Based on such reasoning, we propose a new empirical metric QICE, which by design can be viewed as PICP with finer granularity, and without uncovered quantile ranges. To compute QICE, we first generate enough $\vy$ samples given each $\vx$, and divide them into $M$ bins with roughly equal sizes. We would obtain the corresponding quantile values at each boundary. In this paper, we set $M=10$, and obtain the following $10$ quantile intervals (QIs) of the generated $\vy$ samples: below the $10^{th}$ percentile, between the $10^{th}$ and $20^{th}$ percentiles, $\dots$, between the $80^{th}$ and $90^{th}$ percentiles, and above the $90^{th}$ percentile. Optimally, when the learned conditional distribution is identical to the true one, given enough samples from both learned and true distribution we shall observe about $10\%$ of true data falling into each of these $10$ QIs.

We define \textit{QICE}
to be the mean absolute error between the proportion of true data contained by each QI and the optimal proportion, which is $1/M$ for all intervals:
\ba{
    \text{QICE}\coloneqq \frac{1}{M}\sum_{m=1}^M\bigg|r_m-\frac{1}{M}\bigg|, 
\text{ where }
    r_m = \frac{1}{N}\sum_{n=1}^N\mathbb{1}_{y_n\geq\hat{y}_n^{\text{low}_m}}\cdot\mathbb{1}_{y_n\leq\hat{y}_n^{\text{high}_m}}. \label{eq:QICE}
}
Intuitively, under optimal scenario with enough samples, we shall obtain a QICE value of $0$. Note that each $r_m$ is indeed the PICP for the corresponding QI with boundaries at $\hat{y}_n^{\text{low}_m}$ and $\hat{y}_n^{\text{high}_m}$. Since the true $\vy$ for each $\vx$ is guaranteed to fall into one of these QIs, we are thus able to overcome the mismatch issue described in the above example for PICP: fewer true instances falling into one QI would result in more instances captured by another QI, thus increasing the absolute error for both~QIs.

QICE is similar to NLL in the sense that it also utilizes the summary statistics of the samples from the learned distribution conditional on each new $\vx$ to empirically evaluate how well the model fits the true data. Meanwhile, it does not assume any parametric form on the conditional distribution, making it a much more generalizable metric to measure the level of distributional match between the learned and the underlying true conditional distributions, especially when the true conditional distribution is known to be multi-modal. We will demonstrate this point through the regression toy examples.

\subsubsection{Toy Examples}
\begin{figure*}[t]
    \makebox[\textwidth][c]{
      \includegraphics[width=0.95\textwidth]{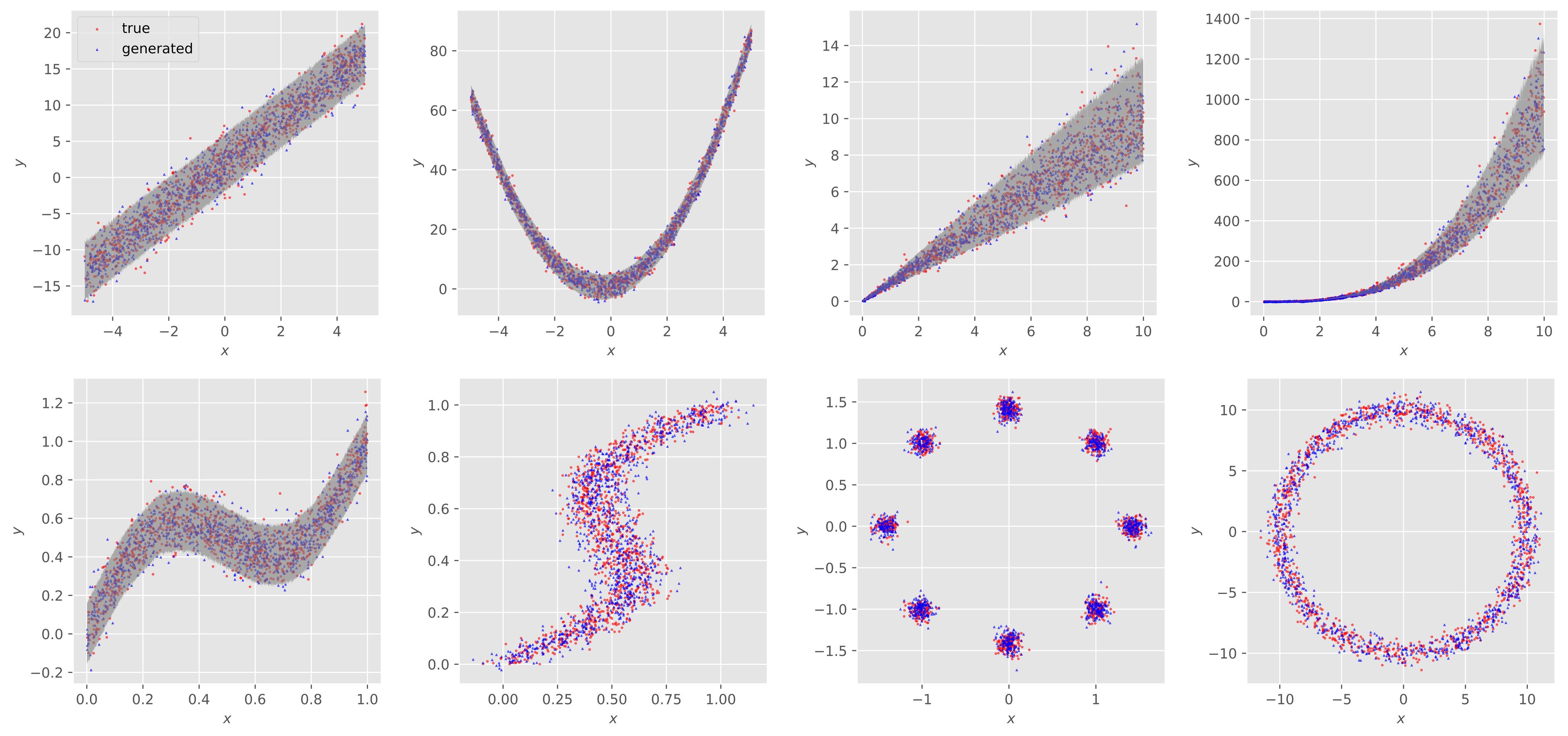}
    }
    \vspace{-5mm}
    \caption{Regression toy example scatter plots. (\textbf{Top}) left to right: linear regression, quadratic regression, log-log linear regression, log-log cubic regression; (\textbf{Bottom}) left to right: sinusoidal regression, inverse sinusoidal regression, 8 Gaussians, full circle.}
    \label{fig:reg_toy_combined}
\end{figure*}

To demonstrate the effectiveness of CARD in regression tasks for not only learning the conditional mean $\mathbb{E}(\vy\given\vx)$, but also recreating the ground truth data generating mechanism, we first apply CARD on $8$ toy examples, whose data generating functions are designed to possess different statistical characteristics: some have a uni-modal symmetric distribution for their error term (linear regression, quadratic regression, sinusoidal regression), others have heteroscedasticity (log-log linear regression, log-log cubic regression) or multi-modality (inverse sinusoidal regression, $8$ Gaussians, full circle). We show that the trained CARD models can generate samples that are visually indistinguishable from the true response variables of the new covariates, as well as quantitatively match the true distribution in terms of some summary statistics. We present the scatter plots of both true and generated data for all $8$ tasks in Figure \ref{fig:reg_toy_combined}. For tasks with uni-modal conditional distribution, we fill the region between the $2.5^{th}$ and $97.5^{th}$ percentile of the generated $\vy$'s. We observe that within each task, the generated samples blend remarkably well with the true test instances, suggesting the capability of reconstructing the underlying data generation mechanism by CARD. A more detailed description of the toy examples, including more quantitative analyses, is presented in Appendix \ref{ssec:reg_toy_details}.

\subsubsection{UCI Regression Tasks}
We continue to investigate our model through experiments on real-world datasets. We adopt the same set of $10$ UCI regression benchmark datasets \citep{ucidataset} as well as the experimental protocol proposed by \citet{pbp} and followed by \citet{mcdropout} and \citet{deepensembles}. The dataset information is provided in Table~\ref{tab:reg_uci_dim}.

We apply multiple train-test splits with $90\%/10\%$ ratio in the same way as \citet{pbp} ($20$ folds for all datasets except $5$ for Protein and $1$ for Year), and report the metrics by their mean and standard deviation across all splits. We compare our method to all aforementioned BNN frameworks: PBP, MC Dropout, and Deep Ensembles, as well as another deep generative model that estimates a conditional distribution sampler, GCDS \citep{jointmatching}. We note that GCDS is related to a concurrent work of \citet{yang2022regularized}, who share a comparable idea but use it in a different application. Following the same paradigm of BNN model assessment, we evaluate the accuracy and predictive uncertainty estimation of CARD by reporting RMSE and NLL. Furthermore, we also report QICE for all methods to evaluate distributional matching. Since this new metric was not applied in previous methods, we re-ran the experiments for all BNNs and obtained comparable or slightly better results in terms of other commonly used metrics reported in their literature. Further details about the experimental setup for these models can be found in Appendix \ref{ssec:uci_baseline_experiment_details}. The experiment results with corresponding metrics are shown in Tables \ref{tab:reg_uci_rmse}, \ref{tab:reg_uci_nll}, and \ref{tab:reg_uci_qice}, with the number of times that each model achieves the best corresponding metric reported in the last row.

\begin{table}[h]
\caption{\label{tab:reg_uci_rmse}RMSE of UCI regression tasks. For both $^1$Kin8nm and $^2$Naval dataset, we multiply the response variable by $100$ to match the scale of others.}\vspace{1mm}
\begin{center}
\resizebox{9cm}{!}{
\begin{tabular}{@{}l|ccccc@{}}
\toprule[1.5pt]
Dataset             &                    &                    & RMSE $\downarrow$               &                    &                    \\
                    & PBP                & MC Dropout         & Deep Ensembles     & GCDS               & CARD (ours)               \\ \midrule
Boston      & $2.89\pm 0.74$ & $3.06\pm 0.96$ & $3.17\pm 1.05$ & $2.75\pm 0.58$ & $\bm{2.61\pm 0.63}$ \\
Concrete            & $5.55\pm 0.46$ & $5.09\pm 0.60$ & $4.91\pm 0.47$ & $5.39\pm 0.55$ & $\bm{4.77\pm 0.46}$ \\
Energy              & $1.58\pm 0.21$ & $1.70\pm 0.22$ & $2.02\pm 0.32$ & $0.64\pm 0.09$ & $\bm{0.52\pm 0.07}$ \\
Kin8nm$^1$              & $9.42\pm 0.29$ & $7.10\pm 0.26$ & $8.65\pm 0.47$ & $8.88\pm 0.42$ & $\bm{6.32\pm 0.18}$ \\
Naval$^2$    & $0.41\pm 0.08$ & $0.08\pm 0.03$ & $0.09\pm 0.01$ & $0.14\pm 0.05$ & $\bm{0.02\pm0.00}$  \\
Power         & $4.10\pm 0.15$ & $4.04\pm 0.14$ & $4.02\pm 0.15$ & $4.11\pm 0.16$ & $\bm{3.93\pm 0.17}$ \\
Protein             & $4.65\pm 0.02$ & $4.16\pm 0.12$ & $4.45\pm 0.02$ & $4.50\pm 0.02$ & $\bm{3.73\pm 0.01}$ \\
Wine                & $0.64\pm 0.04$ & $\bm{0.62\pm 0.04}$ & $0.63\pm 0.04$ & $0.66\pm 0.04$ & $0.63\pm 0.04$ \\
Yacht               & $0.88\pm 0.22$ & $0.84\pm 0.27$ & $1.19\pm 0.49$ & $0.79\pm 0.26$ & $\bm{0.65\pm 0.25}$ \\
Year & $8.86\pm$ NA      & $8.77\pm$ NA            & $8.79\pm$ NA      & $9.20\pm$ NA      & $\bm{8.70\pm}$ NA      \\ \midrule
\# best & $0$      & $1$            & $0$      & $0$      & $\bm{9}$      \\ \bottomrule[1.5pt]
\end{tabular}
}
\end{center}
\vspace{1mm}
\begin{center}
\caption{\label{tab:reg_uci_nll}NLL of UCI regression tasks.}\vspace{1mm}
\resizebox{9cm}{!}{
\begin{tabular}{@{}l|ccccc@{}}
\toprule[1.5pt]
Dataset  &                     &                     & NLL $\downarrow$                &                     &                     \\
         & PBP                 & MC Dropout          & Deep Ensembles      & GCDS                & CARD (ours)               \\ \midrule
Boston   & $2.53\pm 0.27$  & $2.46\pm 0.12$  & $2.35\pm 0.16$  & $18.66\pm 8.92$ & $\bm{2.35\pm 0.12}$  \\
Concrete & $3.19\pm 0.05$  & $3.21\pm 0.18$  & $\bm{2.93\pm 0.12}$  & $13.64\pm 6.88$ & $2.96\pm 0.09$  \\
Energy   & $2.05\pm 0.05$ & $1.50\pm 0.11$  & $1.40\pm 0.27$  & $1.46\pm 0.72$  & $\bm{1.04\pm 0.06}$  \\
Kin8nm   & $-0.83\pm 0.02$ & $-1.14\pm 0.05$ & $-1.06\pm 0.02$ & $-0.38\pm 0.36$ & $\bm{-1.32\pm 0.02}$ \\
Naval    & $-3.97\pm 0.10$  & $-4.45\pm 0.38$ & $-5.94\pm 0.10$ & $-5.06\pm 0.48$ & $\bm{-7.54\pm 0.05}$ \\
Power    & $2.92\pm 0.02$  & $2.90\pm 0.03$ & $2.89\pm 0.02$  & $2.83\pm 0.06$  & $\bm{2.82\pm 0.02}$  \\
Protein  & $3.05\pm 0.00$  & $2.80\pm 0.08$  & $2.89\pm 0.02$  & $2.81\pm 0.09$  & $\bm{2.49\pm 0.03}$  \\
Wine     & $1.03\pm 0.03$  & $0.93\pm 0.06$  & $0.96\pm 0.06$  & $6.52\pm 21.86$ & $\bm{0.92\pm 0.05}$  \\
Yacht    & $1.58\pm 0.08$  & $1.73\pm 0.22$  & $1.11\pm 0.18$  & $\bm{0.61\pm 0.34}$  & $0.90\pm 0.08$  \\
Year     & $3.69\pm$ NA       & $3.42\pm$ NA             & $3.44\pm$ NA       & $3.43\pm$ NA       & $\bm{3.34\pm}$ NA       \\ \midrule
\# best & $0$      & $0$            & $1$      & $1$      & $\bm{8}$      \\ \bottomrule[1.5pt]
\end{tabular}
}
\end{center}
\vspace{1mm}
\begin{center}
\caption{\label{tab:reg_uci_qice}QICE (in $\%$) of UCI regression tasks.}\vspace{1mm}
\resizebox{9cm}{!}{
\begin{tabular}{@{}l|ccccc@{}}
\toprule[1.5pt]
Dataset  &                     &                     & QICE $\downarrow$                &                     &                     \\
         & PBP                 & MC Dropout          & Deep Ensembles      & GCDS                & CARD (ours)                \\ \midrule
Boston   & $3.50\pm 0.88$      & $3.82\pm 0.82$ & $\bm{3.37\pm 0.00}$     & $11.73\pm 1.05$ & $3.45\pm 0.83$     \\
Concrete & $2.52\pm 0.60$      & $4.17\pm 1.06$ & $2.68\pm 0.64$     & $10.49\pm 1.01$ & $\bm{2.30\pm 0.66}$ \\
Energy   & $6.54\pm 0.90$      & $5.22\pm 1.02$ & $\bm{3.62\pm 0.58}$ & $7.41\pm 2.19$ & $4.91\pm 0.94$     \\
Kin8nm   & $1.31\pm 0.25$      & $1.50\pm 0.32$ & $1.17\pm 0.22$     & $7.73\pm 0.80$ & $\bm{0.92\pm 0.25}$ \\
Naval    & $4.06\pm 1.25$ & $12.50\pm 1.95$ & $6.64\pm 0.60$     & $5.76\pm 2.25$ & $\bm{0.80\pm 0.21}$     \\
Power    & $\bm{0.82\pm 0.19}$  & $1.32\pm 0.37$ & $1.09\pm 0.26$     & $1.77\pm 0.33$ & $0.92\pm 0.21$ \\
Protein  & $1.69\pm 0.09$      & $2.82\pm 0.41$ & $2.17\pm 0.16$     & $2.33\pm 0.18$ & $\bm{0.71\pm 0.11}$  \\
Wine     & $\bm{2.22\pm 0.64}$  & $2.79\pm 0.56$ & $2.37\pm 0.63$     & $3.13\pm 0.79$ & $3.39\pm 0.69$     \\
Yacht    & $6.93\pm 1.74$      & $10.33\pm 1.34$ & $7.22\pm 1.41$     & $\bm{5.01\pm 1.02}$ & $8.03\pm 1.17$     \\
Year     & $2.96\pm$ NA           & $2.43\pm$ NA            & $2.56\pm$ NA          & $1.61\pm$ NA      & $\bm{0.53\pm}$ NA     \\ \midrule
\# best & $2$      & $0$            & $2$      & $1$      & $\bm{5}$      \\ \bottomrule[1.5pt]
\end{tabular}
}
\end{center}
\vspace{-3mm}
\end{table}

We observe that CARD outperforms existing methods, often by a considerable margin (especially on larger datasets), in all metrics for most of the datasets, and is competitive with the best method for the remaining ones: we obtain state-of-the-art results in $9$ out of $10$ datasets in terms of RMSE, $8$ out of $10$ for NLL, and $5$ out of $10$ for QICE. It is worth noting that although we do not explicitly optimize our model by MSE or by NLL, we still obtain better results than models trained with these objectives.

\subsection{Classification}
Similar to \citet{deepensembles}, our motivation for classification is not to achieve state-of-the-art performance in terms of mean accuracy on the benchmark datasets, which is strongly related to network architecture design. Our goal is two-fold:
\begin{enumerate}
    \item We aim to solve classification problems via a generative model, emphasizing its capability to improve the performance of a base classifier with deterministic outputs in terms of accuracy;
    \item We intend to provide an alternative sense of uncertainty, by introducing the idea of model confidence at the instance level, \textit{i.e.}, how sure the model is about \textit{each} of its predictions, through the stochasticity of outputs from a generative model.
\end{enumerate}

As another type of supervised learning problems, classification is different from regression mainly for the response variable being discrete class labels instead of continuous values. 
The conventional operation is to cast the classifier output as a point estimate, with a value between $0$ and $1$. Such design is intended for prediction interpretability: since humans already have a cognitive intuition for probabilities \citep{probinterpret}, the output from a classification model is intended to convey a sense of likelihood for a particular class label. In other words, the predicted probability should reflect its confidence, \textit{i.e.}, a level of certainty, in predicting such a label. \citet{calibration} provide the following example of a good classifier, whose output aligns with human intuition for probabilities: if the model outputs a probability prediction of $0.8$, we hope it indicates that the model is $80\%$ sure that its prediction is correct; given $100$ predictions of $0.8$, one shall expect roughly $80$ of them are correct.

In that sense, a good classification algorithm not only can predict the correct label, but also can reflect the true correctness likelihood through its probability predictions, \textit{i.e.}, providing calibrated confidence \citep{calibration}. To evaluate the level of miscalibration by a model, metrics like Expected Calibration Error~(ECE) and Maximum Calibration Error~(MCE) \citep{ece} have been adopted in recent literature \citep{freqbnn, tractablefuncvi} for image classification tasks, and calibration methods like Platt scaling and isotonic regression have been developed to improve such alignment \citep{calibration}. 

Note that these methods are all based on point estimate predictions by the classifier. Furthermore, these alignment metrics can only be computed at a subgroup level in practice, instead of at the instance level. In other words, one may not be able to make the claim with the existing classification framework such that \textit{given a particular test instance}, how confident the classifier is in its prediction to be correct. We discuss our analysis in ECE with more details in Appendix \ref{ssec:ece_discussion}, which may help justify our motivation in introducing an alternative way to measure model confidence at the level of \textit{individual test instances} in Section \ref{sssec:instance_level_conf}.

\subsubsection{Predict with Instance Level Model Confidence via Generative Models}\label{sssec:instance_level_conf}
We propose the following framework to assess model confidence for its predictions \textit{at the instance level}: for each test instance, we first sample $N$ class prototype reconstructions by CARD through the classification version of Algorithm \ref{alg:reg_form_1_inf}, and then perform the following computations:
\begin{enumerate}
    \item We directly calculate the prediction interval width (PIW) between the $2.5^{th}$ and $97.5^{th}$ percentiles of the $N$ reconstructed values for all classes, \textit{i.e.}, with $C$ different classes in total, we would obtain $C$ PIWs for each instance;
    \item We then convert the samples into probability space with Eq. (\ref{eq:cls_output_conversion}), and apply paired two-sample $t$-test as an uncertainty estimation method proposed in \citet{cdropout}: we obtain the most and second most predicted classes for each instance, and test whether the difference in their mean predicted probability is statistically significant.
\end{enumerate}

This framework would require the classifier to not produce the exact same output each time, since the goal is to construct prediction intervals for each of the class labels. Therefore, the class of generative models is a preferable modeling choice due to its ability to produce stochastic outputs, instead of just a point estimate by traditional classifiers.

In practice, we view each one-hot label as a class prototype in real continuous space (introduced in Section \ref{ssec:card_cls}), and we use a generative model to reconstruct this prototype in a stochastic fashion. The intuition is that if the classifier is sure about the class that a particular instance belongs to, it would precisely reconstruct the original prototype vector without much uncertainty; otherwise, different class prototype reconstructions of the same test instance tend to have more variations: under the context of denoising diffusion models, given different samples from the prior distribution at timestep~$T$, the label reconstructions would appear rather different from each other.

\subsubsection{Classification with Model Confidence on CIFAR-10 Dataset}\label{sssec:cifar10_cls}
We demonstrate our experimental results on the CIFAR-10 dataset. We first contextualize the performance of CARD in conventional metrics including accuracy and NLL with other BNNs in ResNet-18 architecture in Table~\ref{tab:cls_acc_comparison}. The metrics of other methods were reported in \citet{newelbo}, a recent work in BNNs that proposes tighter ELBOs to improve variational inference performance and prior hyperparameter optimization. Following the recipe in Section~\ref{ssec:card_cls}, we first pre-train a deterministic classifier with the same ResNet-18 architecture, and achieve a test accuracy of $90.39\%$, with which we proceed to train CARD. We then obtain our instance prediction through majority vote, \textit{i.e.}, the most predicted class label among its $N$ samples for each image input, and achieve an improved test accuracy with a mean of $90.93\%$ across $10$ runs, showing its ability to improve test accuracy from the base classifier. Our NLL result is competitive among the best ones, even though the model is not optimized with a cross-entropy objective function, as we assume the class labels to be in the real continuous space.

\begin{table}[ht]
\caption{\label{tab:cls_acc_comparison}Comparison of accuracy (in $\%$) and NLL for CIFAR-10 classification with other BNNs.}
\begin{center}
\resizebox{\textwidth}{!}{
\begin{tabular}{@{}l|cccccccccccc@{}}
\toprule[1.5pt]
Model         & CMV-MF-VI & CM-MF-VI  & CV-MF-VI & MF-VI     & MC Dropout       & MAP & CARD \\ \midrule
Accuracy & $86.25\pm 0.06$ & $86.66\pm 0.24$ & $79.78\pm 0.30$ & $77.08\pm 1.14$    & $83.64\pm 0.28$ & $84.69\pm 0.35$ & $\bm{90.93\pm 0.02}$ \\ \midrule
NLL & $0.41\pm 0.00$ & $\bm{0.39\pm 0.00}$ & $0.59\pm 0.00$ & $0.68\pm 0.02$    & $0.49\pm 0.00$ & $0.93\pm 0.02$ & $0.46\pm 0.00$ \\ \bottomrule[1.5pt]
\end{tabular}}
\end{center}
\end{table}

We now present the results from one model run with the proposed framework in Section \ref{sssec:instance_level_conf} for evaluating the instance-level prediction confidence. After obtaining the PIW and paired two-sample $t$-test ($\alpha=0.05$) result from each test instance, we first split the test instances into two groups by the correctness of majority-vote predictions, then we obtain only the PIW corresponding to the true class for each instance, and compute the mean PIW of the true class within each group. In addition, we split the test instances by $t$-test rejection status, and compute the mean accuracy in each group. We report the results from these two grouping procedures in Table \ref{tab:cls_piw_ttest}, where the metrics are computed across all test instances and at the level of each true class~label.

\begin{table}[ht]
\caption{\label{tab:cls_piw_ttest}PIW (multiplied by $100$) and $t$-test results for CIFAR-10 classification task.}\vspace{1mm}
\begin{center}
\resizebox{8.5cm}{!}{
\begin{tabular}{@{}l|c|cc|cc@{}}
\toprule
Class & Accuracy  &       \multicolumn{2}{c|}{PIW}           &   \multicolumn{2}{c}{Accuracy by $t$-test Status}                \\
      &           & Correct  & Incorrect & Rejected  & Not-Rejected (Count)   \\ \midrule
All   & $90.95\%$ & $2.37$ & $21.52$  & $91.25\%$ & $42.86\%$ (63) \\ \midrule
1     & $91.00\%$ & $3.28$ & $18.83$  & $91.51\%$ & $45.45\%$ (11)   \\
2     & $96.00\%$ & $0.55$ & $29.27$  & $96.19\%$ & $33.33\%$ (3)    \\
3     & $87.30\%$ & $2.65$ & $24.40$  & $87.55\%$ & $25.00\%$ (4) \\
4     & $81.90\%$ & $5.48$ & $21.45$  & $82.10\%$ & $63.64\%$ (11)  \\
5     & $93.30\%$ & $2.41$ & $30.02$  & $93.67\%$ & $20.00\%$ (5) \\
6     & $84.70\%$ & $4.16$ & $19.57$  & $85.21\%$ & $46.15\%$ (13)         \\
7     & $94.20\%$ & $1.84$ & $26.01$  & $94.38\%$ & $33.33\%$ (3) \\
8     & $92.80\%$ & $1.96$ & $19.35$  & $93.07\%$ & $25.00\%$ (4)   \\
9     & $95.30\%$ & $0.56$ & $15.75$  & $95.49\%$ & $33.33\%$ (3)         \\
10    & $93.00\%$ & $1.50$ & $14.04$  & $93.26\%$ & $50.00\%$ (6)     \\ \bottomrule[1.5pt]
\end{tabular}
}
\end{center}
\end{table}

We observe from Table \ref{tab:cls_piw_ttest} that under the scope of the entire test set, the mean PIW of the true class label among the correct predictions is narrower than that of the incorrect predictions by an order of magnitude, indicating that when CARD is making correct predictions, its class label reconstructions have much smaller variations. We may interpret such results as that CARD can reveal what it does not know through the relativity in reconstruction variations. Furthermore, when comparing the mean PIWs across different classes, we observe that the class with a higher prediction accuracy tends to have a sharper contrast in true label PIW between correct and incorrect predictions; additionally, the PIW values of both correct and incorrect predictions tend to be larger in a less accurate class. Meanwhile, it is worth noting that if we predict the class label by the one with the narrowest PIW for each instance, we can already obtain a test accuracy of $87.84\%$, suggesting a strong correlation between the prediction correctness and instance-level model confidence (in terms of label reconstruction variability). Moreover, we observe that the accuracy of test instances rejected by the $t$-test is much higher than that of the not-rejected ones, both across the entire test set and within each class.

We point out that these metrics can reflect how sure CARD is about the correctness of its predictions, and can thus be used as an important indicator of whether the model prediction \textit{of each instance} can be trusted or not. Therefore, it has the potential to be further applied in the human-machine collaboration domain \citep{humanai1, humanai2, humanai3, humanai4}, such that one can apply such uncertainty measurement to decide if we can directly accept the model prediction, or we need to allocate the instance to humans for further evaluation.

\section{Conclusion}
In this paper, we propose Classification And Regression Diffusion (CARD) models, a class of conditional generative models that approaches supervised learning problems from a conditional generation perspective. Without training with objectives directly related to the evaluation metrics, we achieve state-of-the-art results on benchmark regression tasks. Furthermore, CARD exhibits  a strong ability to represent the conditional distribution with multiple density modes. We also propose a new metric Quantile Interval Coverage Error (QICE), which can be viewed as a generalized version of negative log-likelihood in evaluating how well the model fits the data. Lastly, we introduce a framework to evaluate prediction uncertainty at the instance level for classification tasks.

\section*{Acknowledgments}
The authors acknowledge the support of NSF IIS 1812699 and 2212418, and the Texas Advanced Computing Center (TACC) for providing HPC
resources that have contributed to the research results reported within this paper.

\bibliography{reference,References052016}

\begin{thebibliography}{71}
\providecommand{\natexlab}[1]{#1}
\providecommand{\url}[1]{\texttt{#1}}
\expandafter\ifx\csname urlstyle\endcsname\relax
  \providecommand{\doi}[1]{doi: #1}\else
  \providecommand{\doi}{doi: \begingroup \urlstyle{rm}\Url}\fi

\bibitem[Ardizzone et~al.(2020)Ardizzone, Mackowiak, Rother, and
  Köthe]{genclf2}
Lynton Ardizzone, Radek Mackowiak, Carsten Rother, and Ullrich Köthe.
\newblock Training normalizing flows with the information bottleneck for
  competitive generative classification.
\newblock In \emph{Proceedings of the 34th Conference on Neural Information
  Processing Systems}, 2020.

\bibitem[Austin et~al.(2021)Austin, Johnson, Ho, Tarlow, and
  Berg]{austin2021structured}
Jacob Austin, Daniel Johnson, Jonathan Ho, Danny Tarlow, and Rianne van~den
  Berg.
\newblock Structured denoising diffusion models in discrete state-spaces.
\newblock In \emph{Proceedings of the 35th Conference on Neural Information
  Processing Systems}, 2021.

\bibitem[Bishop(1994)]{mdn}
Christopher Bishop.
\newblock Mixture density networks.
\newblock In \emph{Aston University Neural Computing Research Group Report},
  1994.

\bibitem[Blundell et~al.(2015)Blundell, Cornebise, Kavukcuoglu, and
  Wierstra]{bbb}
Charles Blundell, Julien Cornebise, Koray Kavukcuoglu, and Daan Wierstra.
\newblock Weight uncertainty in neural network.
\newblock In \emph{Proceedings of the 32nd International Conference on Machine
  Learning}. PMLR, 2015.

\bibitem[Brier(1950)]{brierscore}
Glenn~W. Brier.
\newblock Verification of forecasts expressed in terms of probability.
\newblock In \emph{Monthly Weather Review}, volume~8, page~1, 1950.

\bibitem[Cosmides and Tooby(1996)]{probinterpret}
Leda Cosmides and John Tooby.
\newblock Are humans good intuitive statisticians after all? {R}ethinking some
  conclusions from the literature on judgment under uncertainty.
\newblock In \emph{cognition}, volume 58(1), pages 1--73, 1996.

\bibitem[Cox(1961)]{ewma}
David~R. Cox.
\newblock Prediction by exponentially weighted moving averages and related
  methods.
\newblock \emph{Journal of the Royal Statistical Society: Series B
  (Methodological)}, 23\penalty0 (2):\penalty0 414--422, 1961.

\bibitem[Dhariwal and Nichol(2021)]{Dhariwal2021DiffusionMB}
Prafulla Dhariwal and Alexander~Quinn Nichol.
\newblock Diffusion models beat {GAN}s on image synthesis.
\newblock In \emph{Proceedings of the 35th Conference on Neural Information
  Processing Systems}, 2021.

\bibitem[Dua and Graff(2017)]{ucidataset}
Dheeru Dua and Casey Graff.
\newblock {UCI} {M}achine {L}earning {R}epository, 2017.
\newblock URL \url{http://archive.ics.uci.edu/ml}.

\bibitem[Fan et~al.(2021)Fan, Zhang, Tanwisuth, Qian, and Zhou]{cdropout}
Xinjie Fan, Shujian Zhang, Korawat Tanwisuth, Xiaoning Qian, and Mingyuan Zhou.
\newblock Contextual dropout: An efficient sample-dependent dropout module.
\newblock In \emph{Proceedings of the 9th International Conference on Learning
  Representations}, 2021.

\bibitem[Feller(1949)]{feller1949theory}
William Feller.
\newblock On the theory of stochastic processes, with particular reference to
  applications.
\newblock In \emph{Proceedings of the 1st Berkeley Symposium on Mathematical
  Statistics and Probability}, pages 403--432. University of California Press,
  1949.

\bibitem[Fetaya et~al.(2020)Fetaya, Jacobsen, Grathwohl, and Zemel]{genclf1}
Ethan Fetaya, Joern-Henrik Jacobsen, Will Grathwohl, and Richard Zemel.
\newblock Understanding the limitations of conditional generative models.
\newblock In \emph{Proceedings of the 8th International Conference on Learning
  Representations}, 2020.

\bibitem[Gal and Ghahramani(2016)]{mcdropout}
Yarin Gal and Zoubin Ghahramani.
\newblock Dropout as a {B}ayesian approximation: Representing model uncertainty
  in deep learning.
\newblock In \emph{Proceedings of the 33rd International Conference on Machine
  Learning}. PMLR, 2016.

\bibitem[Gal et~al.(2017)Gal, Hron, and Kendall]{concretedropout}
Yarin Gal, Jiri Hron, and Alex Kendall.
\newblock Concrete dropout.
\newblock In \emph{Proceedings of the 31st Conference on Neural Information
  Processing Systems}, 2017.

\bibitem[Gao et~al.(2021)Gao, Saar-Tsechansky, De-Arteaga, Han, Lee, and
  Lease]{humanai4}
Ruijiang Gao, Maytal Saar-Tsechansky, Maria De-Arteaga, Ligong Han, Min~Kyung
  Lee, and Matthew Lease.
\newblock {H}uman-{AI} collaboration with bandit feedback.
\newblock In \emph{Proceedings of the 30th International Joint Conferences on
  Artificial Intelligence}, 2021.

\bibitem[Garnelo et~al.(2018{\natexlab{a}})Garnelo, Rosenbaum, Maddison,
  Ramalho, Saxton, Shanahan, Teh, Rezende, and Eslami]{cnp}
Marta Garnelo, Dan Rosenbaum, Chris~J. Maddison, Tiago Ramalho, David Saxton,
  Murray Shanahan, Yee~Whye Teh, Danilo~J. Rezende, and S.~M.~Ali Eslami.
\newblock Conditional neural processes.
\newblock In \emph{Proceedings of the 35th International Conference on Machine
  Learning}, 2018{\natexlab{a}}.

\bibitem[Garnelo et~al.(2018{\natexlab{b}})Garnelo, Schwarz, Rosenbaum, Viola,
  Rezende, Eslami, and Teh]{np}
Marta Garnelo, Jonathan Schwarz, Dan Rosenbaum, Fabio Viola, Danilo~J. Rezende,
  S.M.~Ali Eslami, and Yee~Whye Teh.
\newblock Neural processes.
\newblock In \emph{ICML 2018 workshop on Theoretical Foundations and
  Applications of Deep Generative Models}, 2018{\natexlab{b}}.

\bibitem[Gordon et~al.(2020)Gordon, Bruinsma, Foong, Requeima, Dubois, and
  Turner]{convcnp}
Jonathan Gordon, Wessel~P. Bruinsma, Andrew Y.~K. Foong, James Requeima, Yann
  Dubois, and Richard~E. Turner.
\newblock Convolutional conditional neural processes.
\newblock In \emph{Proceedings of the 8th International Conference on Learning
  Representations}, 2020.

\bibitem[Guo et~al.(2017)Guo, Pleiss, Sun, and Weinberger]{calibration}
Chuan Guo, Geoff Pleiss, Yu~Sun, and Kilian~Q. Weinberger.
\newblock On calibration of modern neural networks.
\newblock In \emph{Proceedings of the 34th International Conference on Machine
  Learning}. PMLR, 2017.

\bibitem[He et~al.(2020)He, Fan, Wu, Xie, and Girshick]{he2020momentum}
Kaiming He, Haoqi Fan, Yuxin Wu, Saining Xie, and Ross Girshick.
\newblock Momentum contrast for unsupervised visual representation learning.
\newblock In \emph{Proceedings of the IEEE/CVF Conference on Computer Vision
  and Pattern Recognition}, pages 9729--9738, 2020.

\bibitem[Hernández-Lobato and Adams(2015)]{pbp}
José~Miguel Hernández-Lobato and Ryan~P. Adams.
\newblock Probabilistic backpropagation for scalable learning of {B}ayesian
  neural networks.
\newblock In \emph{Proceedings of the 32nd International Conference on Machine
  Learning}. PMLR, 2015.

\bibitem[Ho et~al.(2020)Ho, Jain, and Abbeel]{ddpm}
Jonathan Ho, Ajay Jain, and Pieter Abbeel.
\newblock Denoising diffusion probabilistic models.
\newblock In \emph{Proceedings of the 34th Conference on Neural Information
  Processing Systems}, 2020.

\bibitem[Hoogeboom et~al.(2021)Hoogeboom, Nielsen, Jaini, Forré, and
  Welling]{multimomialdiffusion}
Emiel Hoogeboom, Didrik Nielsen, Priyank Jaini, Patrick Forré, and Max
  Welling.
\newblock Argmax flows and multinomial diffusion: Learning categorical
  distributions.
\newblock In \emph{Proceedings of the 35th Conference on Neural Information
  Processing Systems}, 2021.

\bibitem[Hyv{\"a}rinen and Dayan(2005)]{hyvarinen2005estimation}
Aapo Hyv{\"a}rinen and Peter Dayan.
\newblock Estimation of non-normalized statistical models by score matching.
\newblock \emph{Journal of Machine Learning Research}, 6\penalty0 (4), 2005.

\bibitem[Joachims(2021)]{nnuncert}
Per Joachims.
\newblock nnuncert: Uncertainty quantification with {BNN}s, 2021.
\newblock URL \url{https://github.com/nnuncert/nnuncert}.

\bibitem[Kawar et~al.(2022)Kawar, Elad, Ermon, and Song]{kawar2022denoising}
Bahjat Kawar, Michael Elad, Stefano Ermon, and Jiaming Song.
\newblock Denoising diffusion restoration models.
\newblock In \emph{ICLR 2022 Workshop on Deep Generative Models for Highly
  Structured Data}, 2022.

\bibitem[Kendall and Gal(2017)]{alex2017uncertainty}
Alex Kendall and Yarin Gal.
\newblock What uncertainties do we need in {B}ayesian deep learning for
  computer vision?
\newblock In \emph{Proceedings of the 31st Conference on Neural Information
  Processing Systems}, 2017.

\bibitem[Kim et~al.(2019)Kim, Mnih, Schwarz, Garnelo, Eslami, Rosenbaum,
  Vinyals, and Teh]{anp}
Hyunjik Kim, Andriy Mnih, Jonathan Schwarz, Marta Garnelo, Ali Eslami, Dan
  Rosenbaum, Oriol Vinyals, and Yee~Whye Teh.
\newblock Attentive neural processes.
\newblock In \emph{Proceedings of the 7th International Conference on Learning
  Representations}, 2019.

\bibitem[Kingma and Ba(2015)]{adam}
Diederik~P. Kingma and Jimmy Ba.
\newblock Adam: A method for stochastic optimization.
\newblock In \emph{Proceedings of the 3rd International Conference on Learning
  Representations}, 2015.

\bibitem[Kingma et~al.(2015)Kingma, Salimans, and Welling]{vdropout}
Durk~P. Kingma, Tim Salimans, and Max Welling.
\newblock Variational dropout and the local reparameterization trick.
\newblock In \emph{Proceedings of the 29th Conference on Neural Information
  Processing Systems}, 2015.

\bibitem[Kristiadi et~al.(2022)Kristiadi, Hein, and Hennig]{freqbnn}
Agustinus Kristiadi, Matthias Hein, and Philipp Hennig.
\newblock Being a bit frequentist improves {B}ayesian neural networks.
\newblock In \emph{Proceedings of the 25th International Conference on
  Artificial Intelligence and Statistics}, 2022.

\bibitem[Lakshminarayanan et~al.(2017)Lakshminarayanan, Pritzel, and
  Blundell]{deepensembles}
Balaji Lakshminarayanan, Alexander Pritzel, and Charles Blundell.
\newblock Simple and scalable predictive uncertainty estimation using deep
  ensembles.
\newblock In \emph{Proceedings of the 31st Conference on Neural Information
  Processing Systems}, 2017.

\bibitem[LeCun et~al.(1998)LeCun, Bottou, Bengio, and Haffner]{lenet}
Yann LeCun, Léon Bottou, Yoshua Bengio, and Patrick Haffner.
\newblock Gradient-based learning applied to document recognition.
\newblock In \emph{Proceedings of the IEEE}, volume~86, pages 2278--2324, 1998.
\newblock \doi{10.1109/5.726791}.

\bibitem[Liang et~al.(2022)Liang, Zhu, Wang, and Yang]{imagenetresnet50_1}
Yuanzhi Liang, Linchao Zhu, Xiaohan Wang, and Yi~Yang.
\newblock A simple episodic linear probe improves visual recognition in the
  wild.
\newblock In \emph{Proceedings of the IEEE/CVF Conference on Computer Vision
  and Pattern Recognition}, 2022.

\bibitem[Liu et~al.(2021)Liu, Zhou, Jiao, and Huang]{wganjointmatching}
Shiao Liu, Xingyu Zhou, Yuling Jiao, and Jian Huang.
\newblock Wasserstein generative learning of conditional distribution.
\newblock \emph{arXiv preprint arXiv:2112.10039}, 2021.

\bibitem[Liu et~al.(2022)Liu, Chen, Atashgahi, Chen, Sokar, Mocanu,
  Pechenizkiy, Wang, and Mocanu]{deepensemblesrecent}
Shiwei Liu, Tianlong Chen, Zahra Atashgahi, Xiaohan Chen, Ghada Sokar, Elena
  Mocanu, Mykola Pechenizkiy, Zhangyang Wang, and Decebal~Constantin Mocanu.
\newblock Deep ensembling with no overhead for either training or testing: The
  all-round blessings of dynamic sparsity.
\newblock In \emph{Proceedings of the 10th International Conference on Learning
  Representations}, 2022.

\bibitem[Loshchilov and Hutter(2017)]{cosinedecay}
Ilya Loshchilov and Frank Hutter.
\newblock {SGDR}: Stochastic gradient descent with warm restarts.
\newblock In \emph{Proceedings of the 5th International Conference on Learning
  Representations}, 2017.

\bibitem[Mackowiak et~al.(2021)Mackowiak, Ardizzone, Köthe, and
  Rother]{genclf3}
Radek Mackowiak, Lynton Ardizzone, Ullrich Köthe, and Carsten Rother.
\newblock Generative classifiers as a basis for trustworthy image
  classification.
\newblock In \emph{Proceedings of the IEEE/CVF Conference on Computer Vision
  and Pattern Recognition}, 2021.

\bibitem[Madras et~al.(2018)Madras, Pitassi, and Zemel]{humanai1}
David Madras, Toniann Pitassi, and Richard Zemel.
\newblock Predict responsibly: Improving fairness and accuracy by learning to
  defer.
\newblock In \emph{Proceedings of the 32nd Conference on Neural Information
  Processing Systems}, 2018.

\bibitem[Mukhoti and Gal(2018)]{pavpu}
Jishnu Mukhoti and Yarin Gal.
\newblock Evaluating {B}ayesian deep learning methods for semantic
  segmentation.
\newblock \emph{arXiv preprint arXiv: 1811.12709}, 2018.

\bibitem[Naeini et~al.(2015)Naeini, Cooper, and Hauskrecht]{ece}
Mahdi~Pakdaman Naeini, Gregory~F. Cooper, and Milos Hauskrecht.
\newblock Obtaining well calibrated probabilities using {B}ayesian binning.
\newblock In \emph{Proceedings of the 29th AAAI Conference on Artificial
  Intelligence}, 2015.

\bibitem[Neal(2011)]{neal2011mcmc}
Radford~M Neal.
\newblock {MCMC} using {H}amiltonian dynamics.
\newblock \emph{Handbook of Markov Chain Monte Carlo}, page 113, 2011.

\bibitem[Nichol et~al.(2022)Nichol, Dhariwal, Ramesh, Shyam, Mishkin, McGrew,
  Sutskever, and Chen]{nichol2021glide}
Alex Nichol, Prafulla Dhariwal, Aditya Ramesh, Pranav Shyam, Pamela Mishkin,
  Bob McGrew, Ilya Sutskever, and Mark Chen.
\newblock {GLIDE}: Towards photorealistic image generation and editing with
  text-guided diffusion models.
\newblock In \emph{Proceedings of the 39th International Conference on Machine
  Learning}. PMLR, 2022.

\bibitem[Pandey et~al.(2022)Pandey, Mukherjee, Rai, and
  Kumar]{pandey2022diffusevae}
Kushagra Pandey, Avideep Mukherjee, Piyush Rai, and Abhishek Kumar.
\newblock {D}iffuse{VAE}: Efficient, controllable and high-fidelity generation
  from low-dimensional latents.
\newblock \emph{Transactions on Machine Learning Research}, 2022.

\bibitem[Paszke et~al.(2019)Paszke, Gross, Massa, Lerer, Bradbury, Chanan,
  Killeen, Lin, Gimelshein, Antiga, Desmaison, Köpf, Yang, DeVito, Raison,
  Tejani, Chilamkurthy, Steiner, Fang, Bai, and Chintala]{pytorch}
Adam Paszke, Sam Gross, Francisco Massa, Adam Lerer, James Bradbury, Gregory
  Chanan, Trevor Killeen, Zeming Lin, Natalia Gimelshein, Luca Antiga, Alban
  Desmaison, Andreas Köpf, Edward Yang, Zach DeVito, Martin Raison, Alykhan
  Tejani, Sasank Chilamkurthy, Benoit Steiner, Lu~Fang, Junjie Bai, and Soumith
  Chintala.
\newblock Py{T}orch: An imperative style, high-performance deep learning
  library.
\newblock In \emph{Proceedings of the 33rd Conference on Neural Information
  Processing Systems}, 2019.

\bibitem[Pham and Le(2021)]{imagenetresnet50_2}
Hieu Pham and Quoc~V. Le.
\newblock Autodropout: Learning dropout patterns to regularize deep networks.
\newblock In \emph{Proceedings of the 35th AAAI Conference on Artificial
  Intelligence}, 2021.

\bibitem[Raghu et~al.(2019)Raghu, Blumer, Corrado, Kleinberg, Obermeyer, and
  Mullainathan]{humanai2}
Maithra Raghu, Katy Blumer, Greg Corrado, Jon Kleinberg, Ziad Obermeyer, and
  Sendhil Mullainathan.
\newblock The algorithmic automation problem: Prediction, triage, and human
  effort.
\newblock \emph{arXiv preprint arXiv:1903.12220}, 2019.

\bibitem[Ramesh et~al.(2022)Ramesh, Dhariwal, Nichol, Chu, and
  Chen]{ramesh2022hierarchical}
Aditya Ramesh, Prafulla Dhariwal, Alex Nichol, Casey Chu, and Mark Chen.
\newblock Hierarchical text-conditional image generation with {CLIP} latents.
\newblock \emph{arXiv preprint arXiv:2204.06125}, 2022.

\bibitem[Reddi et~al.(2018)Reddi, Kale, and Kumar]{amsgrad}
Sashank~J. Reddi, Satyen Kale, and Sanjiv Kumar.
\newblock On the convergence of {A}dam and beyond.
\newblock In \emph{Proceedings of the 6th International Conference on Learning
  Representations}, 2018.

\bibitem[Ren et~al.(2019)Ren, Zhao, and Ermon]{antitheticsampling}
Hongyu Ren, Shengjia Zhao, and Stefano Ermon.
\newblock Adaptive antithetic sampling for variance reduction.
\newblock In \emph{Proceedings of the 36th International Conference on Machine
  Learning}. PMLR, 2019.

\bibitem[Revow et~al.(1996)Revow, Williams, and Hinton]{genclfearlywork}
Michael Revow, Christopher~KI Williams, and Geoffrey~E Hinton.
\newblock Using generative models for handwritten digit recognition.
\newblock \emph{IEEE Transactions on Pattern Analysis and Machine
  Intelligence}, 18\penalty0 (6):\penalty0 592--606, 1996.

\bibitem[Rudner et~al.(2021)Rudner, Chen, Teh, and Gal]{tractablefuncvi}
Tim G.~J. Rudner, Zonghao Chen, Yee~Whye Teh, and Yarin Gal.
\newblock Tractable function-space variational inference in {B}ayesian neural
  networks.
\newblock In \emph{ICML 2021 Workshop on Uncertainty and Robustness in Deep
  Learning}, 2021.

\bibitem[Sohl-Dickstein et~al.(2015)Sohl-Dickstein, Weiss, Maheswaranathan, and
  Ganguli]{diffusion}
Jascha Sohl-Dickstein, Eric~A. Weiss, Niru Maheswaranathan, and Surya Ganguli.
\newblock Deep unsupervised learning using nonequilibrium thermodynamics.
\newblock In \emph{Proceedings of the 32nd International Conference on Machine
  Learning}. PMLR, 2015.

\bibitem[Song et~al.(2021{\natexlab{a}})Song, Meng, and Ermon]{ddim}
Jiaming Song, Chenlin Meng, and Stefano Ermon.
\newblock Denoising diffusion implicit models.
\newblock In \emph{Proceedings of the 9th International Conference on Learning
  Representations}, 2021{\natexlab{a}}.

\bibitem[Song and Ermon(2019)]{scorematching}
Yang Song and Stefano Ermon.
\newblock Generative modeling by estimating gradients of the data distribution.
\newblock In \emph{Proceedings of the 33rd Conference on Neural Information
  Processing Systems}, 2019.

\bibitem[Song and Ermon(2020)]{improvedscore}
Yang Song and Stefano Ermon.
\newblock Improved techniques for training score-based generative models.
\newblock In \emph{Proceedings of the 34th Conference on Neural Information
  Processing Systems}, 2020.

\bibitem[Song et~al.(2021{\natexlab{b}})Song, Durkan, Murray, and
  Ermon]{song2021maximum}
Yang Song, Conor Durkan, Iain Murray, and Stefano Ermon.
\newblock Maximum likelihood training of score-based diffusion models.
\newblock In \emph{Proceedings of the 35th Conference on Neural Information
  Processing Systems}, 2021{\natexlab{b}}.

\bibitem[Song et~al.(2021{\natexlab{c}})Song, Sohl-Dickstein, Kingma, Kumar,
  Ermon, and Poole]{song2021scorebased}
Yang Song, Jascha Sohl-Dickstein, Diederik~P Kingma, Abhishek Kumar, Stefano
  Ermon, and Ben Poole.
\newblock Score-based generative modeling through stochastic differential
  equations.
\newblock In \emph{Proceedings of the 9th International Conference on Learning
  Representations}, 2021{\natexlab{c}}.

\bibitem[Srivastava et~al.(2014)Srivastava, Hinton, Krizhevsky, Sutskever, and
  Salakhutdinov]{gausdropout}
Nitish Srivastava, Geoffrey Hinton, Alex Krizhevsky, Ilya Sutskever, and Ruslan
  Salakhutdinov.
\newblock Dropout: A simple way to prevent neural networks from overfitting.
\newblock \emph{Journal of Machine Learning Research}, 15\penalty0
  (1):\penalty0 1929--1958, 2014.

\bibitem[Tomczak et~al.(2021)Tomczak, Swaroop, Foong, and Turner]{newelbo}
Marcin~B. Tomczak, Siddharth Swaroop, Andrew Y.~K. Foong, and Richard~E.
  Turner.
\newblock Collapsed variational bounds for {B}ayesian neural networks.
\newblock In \emph{Proceedings of the 35th Conference on Neural Information
  Processing Systems}, 2021.

\bibitem[Vincent(2011)]{vincent2011connection}
Pascal Vincent.
\newblock A connection between score matching and denoising autoencoders.
\newblock \emph{Neural computation}, 23\penalty0 (7):\penalty0 1661--1674,
  2011.

\bibitem[Wang and Zhou(2020)]{wang2020thompson}
Zhendong Wang and Mingyuan Zhou.
\newblock Thompson sampling via local uncertainty.
\newblock In \emph{Proceedings of the 37th International Conference on Machine
  Learning}. PMLR, 2020.

\bibitem[Welling and Teh(2011)]{welling2011bayesian}
Max Welling and Yee~W Teh.
\newblock {B}ayesian learning via stochastic gradient {L}angevin dynamics.
\newblock In \emph{Proceedings of the 28th International Conference on Machine
  Learning}. PMLR, 2011.

\bibitem[Wilder et~al.(2020)Wilder, Horvitz, and Kamar]{humanai3}
Bryan Wilder, Eric Horvitz, and Ece Kamar.
\newblock Learning to complement humans.
\newblock In \emph{Proceedings of the 29th International Joint Conferences on
  Artificial Intelligence}, 2020.

\bibitem[Xiao et~al.(2022)Xiao, Kreis, and Vahdat]{xiao2021tackling}
Zhisheng Xiao, Karsten Kreis, and Arash Vahdat.
\newblock Tackling the generative learning trilemma with denoising diffusion
  {GANs}.
\newblock In \emph{Proceedings of the 10th International Conference on Learning
  Representations}, 2022.

\bibitem[Yang et~al.(2022)Yang, Wang, Zheng, Feng, and
  Zhou]{yang2022regularized}
Shentao Yang, Zhendong Wang, Huangjie Zheng, Yihao Feng, and Mingyuan Zhou.
\newblock A regularized implicit policy for offline reinforcement learning.
\newblock \emph{arXiv preprint arXiv:2202.09673}, 2022.

\bibitem[Yao et~al.(2019)Yao, Pan, Ghosh, and Doshi-Velez]{bnnquality}
Jiayu Yao, Weiwei Pan, Soumya Ghosh, and Finale Doshi-Velez.
\newblock Quality of uncertainty quantification for {B}ayesian neural network
  inference.
\newblock In \emph{ICML 2019 Workshop on Uncertainty and Robustness in Deep
  Learning}, 2019.

\bibitem[Zheng et~al.(2021)Zheng, Chen, Yao, Yang, Li, Zhang, Zhang, Tsang,
  Zhou, and Zhou]{zheng2021contrastive}
Huangjie Zheng, Xu~Chen, Jiangchao Yao, Hongxia Yang, Chunyuan Li, Ya~Zhang,
  Hao Zhang, Ivor Tsang, Jingren Zhou, and Mingyuan Zhou.
\newblock Contrastive attraction and contrastive repulsion for representation
  learning.
\newblock \emph{arXiv preprint arXiv:2105.03746}, 2021.

\bibitem[Zheng et~al.(2022)Zheng, He, Chen, and Zhou]{truncdiff}
Huangjie Zheng, Pengcheng He, Weizhu Chen, and Mingyuan Zhou.
\newblock Truncated diffusion probabilistic models and diffusion-based
  adversarial auto-encoders.
\newblock \emph{arXiv preprint arXiv:2202.09671}, 2022.

\bibitem[Zhou et~al.(2021)Zhou, Jiao, Liu, and Huang]{jointmatching}
Xingyu Zhou, Yuling Jiao, Jin Liu, and Jian Huang.
\newblock A deep generative approach to conditional sampling.
\newblock \emph{Journal of the American Statistical Association}, pages 1--28,
  2021.

\bibitem[Zimmermann et~al.(2021)Zimmermann, Schott, Song, Dunn, and
  Klindt]{genclf}
Roland~S. Zimmermann, Lukas Schott, Yang Song, Benjamin~A. Dunn, and David~A.
  Klindt.
\newblock Score-based generative classifiers.
\newblock In \emph{NeurIPS 2021 Workshop on Deep Generative Models and
  Downstream Applications}, 2021.

\end{thebibliography}
\bibliographystyle{plainnat}
\clearpage

\newpage
\appendix

\section{Derivations and Additional Experiment Results}

\subsection{Derivations}\label{ssec:fwd_process_post_derivation}
\paragraph{Derivation for Forward Process Posteriors:}
In this section, we derive the mean and variance of the forward process posteriors $q(\vy_{t-1}\given\vy_t, \vy_0, \vx)$ in Eq. (\ref{eq:form_1_posterior}), which are tractable when conditioned on $\vy_0$:
\ba{
    &q(\vy_{t-1}\given\vy_t, \vy_0, \vx)
    \propto \underbrace{q\big(\vy_t\given\vy_{t-1}, f_{\phi}(\vx)\big)}_{\text{Eq. }(\ref{eq:form_1_forward_one_step})}\underbrace{q\big(\vy_{t-1}\given\vy_0, f_{\phi}(\vx)\big)}_{\text{Eq. }(\ref{eq:form_1_forward_t_steps})} \\
    &\propto \exp\Bigg(-\frac{1}{2}\bigg(\frac{\big(\vy_t-(1-\sqrt{\alpha_t})f_{\phi}(\vx)-\sqrt{\alpha_t}\vy_{t-1}\big)^2}{\beta_t} \notag \\
    &+\frac{\big(\vy_{t-1}-\sqrt{\bar{\alpha}_{t-1}}\vy_0-(1-\sqrt{\bar{\alpha}_{t-1}})f_{\phi}(\vx)\big)^2}{1-\bar{\alpha}_{t-1}}\bigg)\Bigg) \\
    &\propto \exp\Bigg(-\frac{1}{2}\bigg(\frac{\alpha_t\vy_{t-1}^2-2\sqrt{\alpha_t}\big(\vy_t-(1-\sqrt{\alpha_t})f_{\phi}(\vx)\big)\vy_{t-1}}{\beta_t} \notag \\ 
    &+\frac{\vy_{t-1}^2-2\big(\sqrt{\bar{\alpha}_{t-1}}\vy_0+(1-\sqrt{\bar{\alpha}_{t-1}})f_{\phi}(\vx)\big)\vy_{t-1}}{1-\bar{\alpha}_{t-1}}\bigg)\Bigg) \\
    &= \exp\Bigg(-\frac{1}{2}\bigg(\Big(\underbrace{\frac{\alpha_t}{\beta_t}+\frac{1}{1-\bar{\alpha}_{t-1}}}_{\circled{1}}\Big)\vy_{t-1}^2 \notag \\
    &-2\Big(\underbrace{\frac{\sqrt{\bar{\alpha}_{t-1}}}{1-\bar{\alpha}_{t-1}}\vy_0+\frac{\sqrt{\alpha_t}}{\beta_t}\vy_t+\big(\frac{\sqrt{\alpha_t}(\sqrt{\alpha_t}-1)}{\beta_t}+\frac{1-\sqrt{\bar{\alpha}_{t-1}}}{1-\bar{\alpha}_{t-1}}\big)f_{\phi}(\vx)}_{\circled{2}}\Big)\vy_{t-1}\bigg)\Bigg), \label{eq:form_1_fwd_posterior}
}
where
\ba{
\circled{1}=\frac{\alpha_t(1-\bar{\alpha}_{t-1})+\beta_t}{\beta_t(1-\bar{\alpha}_{t-1})}=\frac{1-\bar{\alpha}_t}{\beta_t(1-\bar{\alpha}_{t-1})},
}
and we have the posterior variance
\ba{
\tilde{\beta}_t=\frac{1}{\circled{1}}=\frac{1-\bar{\alpha}_{t-1}}{1-\bar{\alpha}_t}\beta_t.
}
Meanwhile, the following coefficients of the terms in the posterior mean through dividing each coefficient in $\circled{2}$ by $\circled{1}$:
\ba{
\gamma_0 = \frac{\sqrt{\bar{\alpha}_{t-1}}}{1-\bar{\alpha}_{t-1}}/\circled{1}=\frac{\sqrt{\bar{\alpha}_{t-1}}}{1-\bar{\alpha}_t}\beta_t,
}
\ba{
\gamma_1 = \frac{\sqrt{\alpha_t}}{\beta_t}/\circled{1}=\frac{1-\bar{\alpha}_{t-1}}{1-\bar{\alpha}_t}\sqrt{\alpha_t},
}
and 
\ba{
\gamma_2 &= \bigg(\frac{\sqrt{\alpha_t}(\sqrt{\alpha_t}-1)}{\beta_t}+\frac{1-\sqrt{\bar{\alpha}_{t-1}}}{1-\bar{\alpha}_{t-1}}\bigg)/\circled{1} \notag \\
&= \frac{\alpha_t-\bar{\alpha}_t-\sqrt{\alpha_t}(1-\bar{\alpha}_{t-1})+\beta_t-\beta_t\sqrt{\bar{\alpha}_{t-1}}}{1-\bar{\alpha}_t} \notag \\
&= 1+\frac{(\sqrt{\bar{\alpha}_t}-1)(\sqrt{\alpha_t}+\sqrt{\bar{\alpha}_{t-1}})}{1-\bar{\alpha}_t},
}
which together give us the posterior mean
\ba{
     \tilde{\vmu}\big(\vy_t, \vy_0, f_{\phi}(\vx)\big) =\gamma_0\vy_0+\gamma_1\vy_t+\gamma_2 f_{\phi}(\vx). \notag
}

\paragraph{Derivation for Forward Process Sampling Distribution with Arbitrary Timesteps:}
For completeness, we include the derivation for the parameters of the forward diffusion process sampling distribution with arbitrary $t$ steps.

The expectation term is based on Eqs. $(36)$--$(38)$ in \citet{pandey2022diffusevae}. From Eq. (\ref{eq:form_1_forward_one_step}), we have that for all $t=1,\dots,T$,
\ba{
\vy_t = \sqrt{1-\beta_t}\vy_{t-1} + (1-\sqrt{1-\beta_t})f_{\phi}(\vx) + \sqrt{\beta_t}\epsilonv, \text{ where } \epsilonv\sim\mathcal{N}(\bm{0}, \mI).
}

Taking expectation of both sides, we have
\ba{
\mathbb{E}(\vy_t) &= \sqrt{1-\beta_t}\mathbb{E}(\vy_{t-1}) + (1-\sqrt{1-\beta_t})f_{\phi}(\vx)\\
&= \sqrt{1-\beta_t}\big(\sqrt{1-\beta_{t-1}}\mathbb{E}(\vy_{t-2}) + (1-\sqrt{1-\beta_{t-1}})f_{\phi}(\vx)\big) + (1-\sqrt{1-\beta_t})f_{\phi}(\vx) \notag \\
&= \sqrt{(1-\beta_t)(1-\beta_{t-1})}\mathbb{E}(\vy_{t-2}) + \big(1-\sqrt{(1-\beta_t)(1-\beta_{t-1})}\big)f_{\phi}(\vx) \\
&\vdots \notag \\
&= \sqrt{\prod_{i=2}^t(1-\beta_i)}\mathbb{E}(\vy_1) + \Bigg(1-\sqrt{\prod_{i=2}^t(1-\beta_i)}\Bigg)f_{\phi}(\vx) \\
&= \sqrt{\prod_{i=2}^t(1-\beta_i)}\big(\sqrt{1-\beta_1}\vy_{0} + (1-\sqrt{1-\beta_1})f_{\phi}(\vx)\big) + \Bigg(1-\sqrt{\prod_{i=2}^t(1-\beta_i)}\Bigg)f_{\phi}(\vx) \\
&= \sqrt{\prod_{i=1}^t(1-\beta_i)}\vy_0 + \Bigg(1-\sqrt{\prod_{i=1}^t(1-\beta_i)}\Bigg)f_{\phi}(\vx) \\
&= \sqrt{\bar{\alpha}_t}\vy_0 + (1-\sqrt{\bar{\alpha}_t})f_{\phi}(\vx).
}

Meanwhile, since the addition of two independent Gaussians with different variances, $\mathcal{N}(\bm{0}, \sigma_a^2\mI)$ and $\mathcal{N}(\bm{0}, \sigma_b^2\mI)$, is distributed as $\mathcal{N}(\bm{0}, (\sigma_a^2+\sigma_b^2)\mI)$, we can derive the variance term accordingly, 
\ba{
\sigma^2(\vy_t)=(1-\bar{\alpha}_t)\mI.
}

\subsection{More In-Depth Discussion on Several Related Works}\label{ssec:related_work_indepth}
In this section, we discuss the relations between CARD and several existing works, which are briefly addressed in Section \ref{sec:relatedwork}, in more depth.

\subsubsection{Comparing CARD with the Neural Processes Family}
In short, CARD models $p(\vy\given\vx,\mathcal{D}_{\text{in}})$, while the Neural Processes Family (NPF) \citep{np, cnp, anp, convcnp} models $p(\vy\given\vx,\mathcal{D}_{\text{out}})$, where $\mathcal{D}_{\text{in}}$ and $\mathcal{D}_{\text{out}}$ represents in-distribution dataset and out-of-distribution dataset, respectively.

Although both classes of methods can be expressed as modeling $p(\vy\given\vx,\mathcal{D})$, CARD assumes such $(\vx,\vy)$ comes from the same data-generating mechanism as the set $\mathcal{D}$, while NPF assumes $(\vx,\vy)$ to be not from the same distribution as $\mathcal{D}$. While CARD fits in the traditional supervised learning setting for in-distribution generalization, NPF is specifically suited for few-shot learning scenarios, where a good model would capture enough pattern from previously seen datasets so that it can generalize well with very limited samples from the new dataset.

Furthermore, both classes of models are capable of generating stochastic outputs, where CARD aims to capture aleatoric uncertainty, which is intrinsic to the data (thus cannot be reduced), while NPF can express epistemic uncertainty as it proposes more diverse functional forms at regions where data is sparse (and such uncertainty would reduce when more data is given). In terms of the conditioning of $\mathcal{D}$, the information of $\mathcal{D}$ is amortized into the network $\epsilonv_{\rvtheta}$ for CARD, while it is included as an explicit representation in the network of NPF that outputs the distribution parameters for $p(\vy\given\vx)$. It is also worth pointing out that CARD does not assume a parametric distributional form for $p(\vy\given\vx, \mathcal{D})$, while NPF assumes a Gaussian distribution, and designs the objective function with such assumption.

The concept and comparison between epistemic and aleatoric uncertainty is more thoroughly discussed by \citet{alex2017uncertainty}, in which we quote, “Out-of-data examples, which can be identified with epistemic uncertainty, cannot be identified with aleatoric uncertainty alone.” We acknowledge that modeling OOD uncertainty is an important topic for regression tasks; however, we design our model to focus on modeling aleatoric uncertainty in this paper. We plan to explore CARD’s ability in extrapolation as part of our future work.

\subsubsection{Comparing CARD with Discrete Diffusion Models}

To construct our framework for classification, we assume the class labels (in terms of one-hot vectors) are from real continuous spaces instead of discrete ones. This assumption enables us to model the forward diffusion process and prior distribution at timestep $T$ with Gaussian distributions, thus all derivations under the regression setting with analytical computation of KL terms, as well as the corresponding algorithms, generalize naturally into the classification settings. The code for training and inference is exactly the same. Discrete Denoising Diffusion Probabilistic Models (D3PMs) \citep{austin2021structured} fit conventional perception in classification tasks naturally by keeping the assumption of a categorical distribution. Therefore, the corresponding evaluation metrics like NLL can directly translate into such a framework --- we believe that by adopting the discrete space assumption, a better NLL metric can be achieved. Meanwhile, it would require a lot more changes to be made from our framework for regression tasks, including the choice of the transition matrix, the incorporation of $\vx$ into the diffusion processes, as well as the addition of the auxiliary loss into the objective function --- all of the above are classification-task-specific settings, and cannot be adopted with the existing framework for regression tasks.

Besides the intention for consistency and generalizability across the two types of supervised learning tasks, we found that our current construction gives reasonable results to access model prediction confidence at the instance level --- by directly using the prediction intervals obtained in the raw continuous space, \textit{i.e.}, before adopting the softmax function for conversion to probability space, we obtain the sharp contrast in true label PIW between correct and incorrect predictions, and can already achieve high accuracy by merely predicting the label with the narrowest PIW for each instance. However, after converting the reconstructed labels to the probability space, the true label PIW contrast is reduced drastically, and the prediction accuracy by the narrowest PIW is similar to a random guess.

To recap, if achieving the best NLL and ECE for classification is the goal, then we think discrete diffusion models like \citet{austin2021structured} could be excellent choices due to their use of the cross-entropy loss that is directed related to NLL and ECE; however, if the main goal is to access the prediction confidence at the instance level, the proposed CARD framework works well, and it would be interesting to make a head-to-head comparison with discrete diffusion-based classification models that yet need to be developed.

\subsubsection{Comparing CARD with Kendall and Gal (2017)}

\citet{alex2017uncertainty} address BNN as an important class of methods for modeling uncertainty. CARD is similar to BNNs in providing stochastic outputs. However, BNNs deliver such stochasticity by modeling \textit{epistemic} uncertainty, the uncertainty over network parameters $\bm{W}$ (by placing a prior distribution over $\bm{W}$) --- this type of uncertainty is a \textit{property of the model}. On the other hand, CARD does not model epistemic uncertainty, as it applies a deterministic deep neural network as its functional form; it is designed to model \textit{aleatoric} uncertainty instead, which is a \textit{property intrinsic to the data}. In Eq. 2 of \citet{alex2017uncertainty}, such aleatoric uncertainty is captured by the last term as $\sigma^2$, which is a constant with respect to the network parameters $\rvtheta$ for the variational distribution of model parameter $\bm{W}$, thus ignored during the optimization of $\rvtheta$. The new method proposed in \citet{alex2017uncertainty} aims to model the aleatoric uncertainty by making $\sigma^2$ as part of the BNN output (Eq. 7); however, note that it still explicitly assumes $p(\vy\given\vx)$ to be a Gaussian distribution, as the objective function is the negative Gaussian log-likelihood, thus its effectiveness in capturing the actual aleatoric uncertainty depends on the validity of such parametric assumption for $p(\vy\given\vx)$.

\subsubsection{Comparing CARD with Score-Based Generative Classifiers}
From a naming perspective, it might be easy to confuse CARD for classification as a type of \textit{generative~classifiers}, as it utilizes a generative model to conduct classification tasks. However, they are two different types of generative models, as generative classifiers model the conditional distribution $p(\vx\given\vy)$, while CARD models a different conditional distribution, \textit{i.e.}, $p(\vy\given\vx)$. In fact, CARD shall be categorized as a type of discriminative classifier, by the definition in \citet{genclf}. Note that although both types of classifiers under image-based tasks would report NLL as one evaluation metric, they are also different, since the NLL for generative classifiers is evaluated in the space transformed from the logit space of $\vx$, while the NLL for discriminative classifiers is computed in the space of $\vy$ as the cross-entropy between the true label and the predicted probability.

\subsection{Classification on FashionMNIST Dataset}\label{ssec:fmnist_cls}
We perform classification on FashionMNIST dataset with CARD, and present the results in a similar fashion as Section \ref{sssec:cifar10_cls}. We first contextualize the performance of CARD through the accuracy of other BNNs with the LeNet CNN \citep{lenet} architecture in Table~\ref{tab:fmnist_cls_acc_comparison}, where the metrics were first reported in \citet{newelbo}. For this dataset, our pre-trained classifier has the same LeNet architecture as the baselines, which achieves a test accuracy of $91.12\%$. CARD improves the mean test accuracy to $91.79\%$.

\begin{table}[ht]
\caption{\label{tab:fmnist_cls_acc_comparison}Comparison of accuracy (in $\%$) on FashionMNIST dataset with other BNNs.}
\begin{center}
\resizebox{\textwidth}{!}{
\begin{tabular}{@{}l|cccccccccc@{}}
\toprule[1.5pt]
Model         & CMV-MF-VI & CM-MF-VI  & CV-MF-VI  & CM-MF-VI OPT & MF-VI     & MAP       & MC Dropout & MF-VI EB  & $f_{\phi}$ (LeNet) & CARD \\ \midrule
Accuracy & $91.10\pm 0.22$ & $90.95\pm 0.31$ & $88.53\pm 0.13$ & $90.67\pm 0.07$    & $87.04\pm 0.28$ & $88.06\pm 0.22$ & $87.99\pm 0.17$  & $87.04\pm 0.08$ & 91.12 & $\bm{91.79\pm 0.09}$ \\ \bottomrule[1.5pt]
\end{tabular}}
\end{center}
\end{table}

We then present Table \ref{tab:fmnist_cls_piw_ttest}, from which we can draw the same conclusions as Table \ref{tab:cls_piw_ttest} on the CIFAR-10 dataset. Note that we set $\alpha=0.01$ for the paired two-sample $t$-test. When making the prediction for each instance merely by the class with the narrowest PIW, we obtain a test accuracy of $89.36\%$.

\begin{table}[ht]
\caption{\label{tab:fmnist_cls_piw_ttest}PIW (multiplied by $100$) and $t$-test results for FashionMNIST classification task.}\vspace{1mm}
\begin{center}
\resizebox{9cm}{!}{
\begin{tabular}{@{}l|c|cc|cc@{}}
\toprule
Class & Accuracy  &       \multicolumn{2}{c|}{PIW}           &   \multicolumn{2}{c}{Accuracy by $t$-test Status}                \\
      &           & Correct  & Incorrect & Rejected  & Not-Rejected (Count)   \\ \midrule
All   & $91.79\%$ & $0.67$ & $3.20$  & $92.07\%$ & $55.84\%$ (77) \\ \midrule
1     & $88.10\%$ & $0.96$ & $3.40$  & $88.45\%$ & $61.54\%$ (13)   \\
2     & $98.50\%$ & $0.39$ & $2.08$  & $98.60\%$ & $66.67\%$ (3)    \\
3     & $87.70\%$ & $0.84$ & $3.42$  & $88.00\%$ & $50.00\%$ (8) \\
4     & $91.10\%$ & $0.76$ & $2.97$  & $91.59\%$ & $53.85\%$ (13)  \\
5     & $87.90\%$ & $0.89$ & $2.91$  & $88.40\%$ & $33.33\%$ (9) \\
6     & $97.20\%$ & $0.41$ & $2.89$  & $97.29\%$ & $66.67\%$ (3)         \\
7     & $74.80\%$ & $1.37$ & $3.26$  & $74.90\%$ & $70.00\%$ (20) \\
8     & $97.40\%$ & $0.49$ & $1.60$  & $97.50\%$ & $50.00\%$ (2)   \\
9     & $98.40\%$ & $0.34$ & $1.93$  & $98.50\%$ & $0.00\%$ (1)         \\
10    & $96.80\%$ & $0.46$ & $5.59$  & $97.09\%$ & $40.00\%$ (5)     \\ \bottomrule[1.5pt]
\end{tabular}
}
\end{center}
\end{table}

\subsection{Test for Normality Assumption of Paired Two-Sample \textit{t}-test}
\begin{figure*}[ht]
    \centering
    \includegraphics[width=\textwidth]{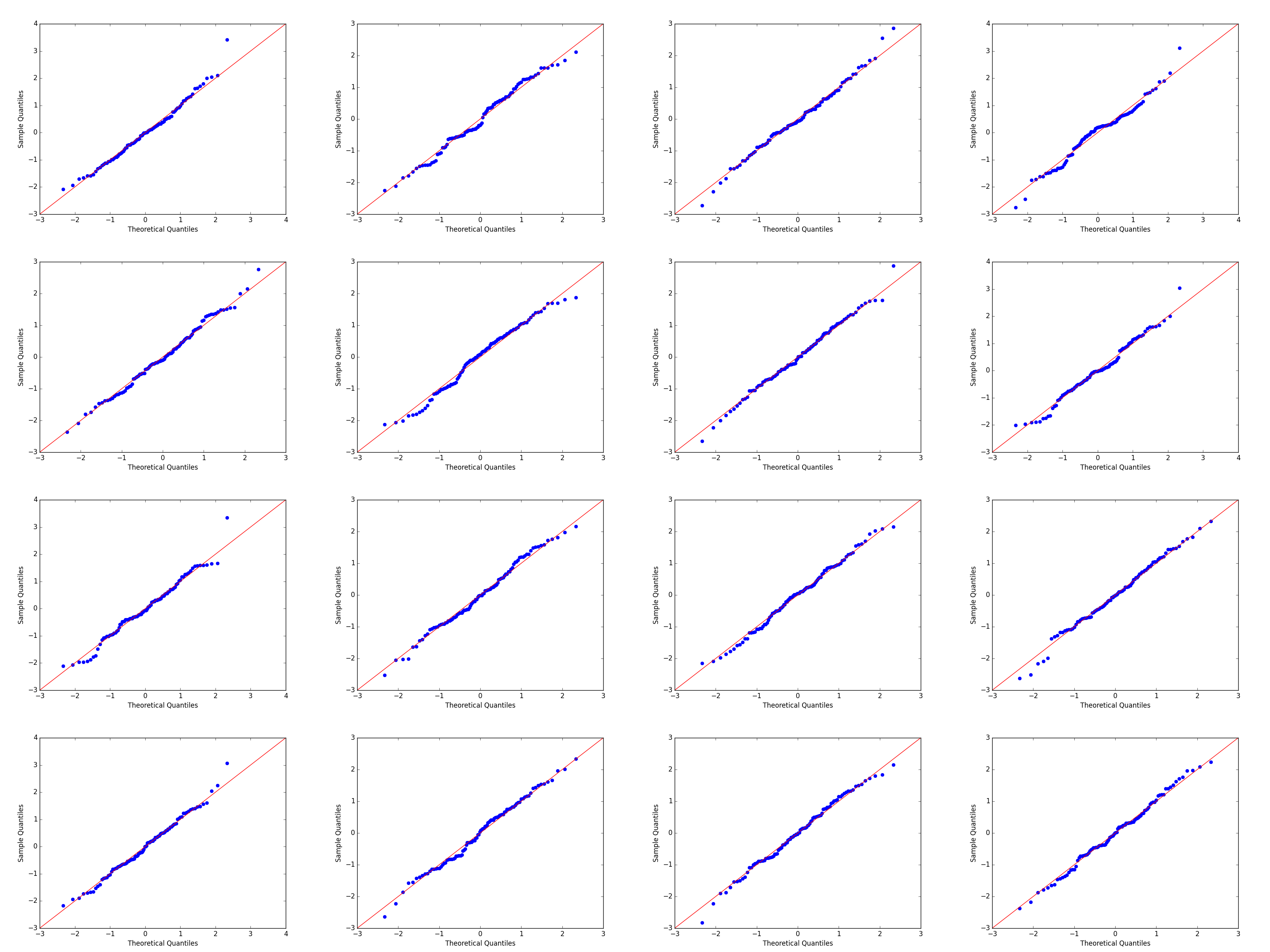}
    \vspace{-5mm}
    \caption{Q-Q plots for the differences in probability between the most and $2^{nd}$ most predicted class.}
    \label{fig:cls_qq_plots}
\end{figure*}

To assess the normality assumption of the paired two-sample $t$-test, we inspect the Q-Q plots of the differences in probability between the most and the second most predicted classes within each test instance. We include $16$ instances in Figure \ref{fig:cls_qq_plots} from CARD predictions on FashionMNIST dataset, each with $100$ samples. We observe that in all plots, the points align closely with the $45$-degree line, indicating that the normality assumption is valid.

\subsection{Patch Accuracy vs Patch Uncertainty (PAvPU)}
Besides the methods of assessing model confidence at the instance level introduced in this paper, we also consider an additional uncertainty evaluation metric in this section, \emph{Patch Accuracy vs Patch Uncertainty} (PAvPU)~\citep{pavpu}, which measures the proportion of predictions that are either correct while the model is confident about them, or incorrect while the model is ambiguous. Based on the $t$-test results, we can easily compute PAvPU, which is defined as
\ba{
\text{PAvPU} \coloneqq \frac{n_{ac}+n_{iu}}{n_{ac}+n_{au}+n_{ic}+n_{iu}}, \label{eq:pavpu}
}
where $n_{ac}, n_{au} ,n_{ic}, n_{iu}$ represent the number of accurate (correct) predictions when the model is certain (confident) about them, accurate but uncertain, inaccurate but certain, as well as inaccurate and uncertain. We apply the $t$-test results as the proxy for the model's confidence level on each of its predictions. A higher value indicates that the model tends to be correct when being confident, and to make mistakes when being vague --- a characteristic that we want our model to possess. On a related note, accuracy can be computed by replacing $n_{iu}$ with $n_{au}$ in Eq. (\ref{eq:pavpu}). In the following section, we provide additional experimental results including this metric.

\subsection{Classification on Noisy MNIST dataset}\label{ssec:noisy_mnist}
To further demonstrate the effectiveness of the CARD model for classification, especially in expressing model confidence, we run additional experiments on Noisy MNIST dataset (adding a Gaussian noise with mean $0$ and variance $1$ to each pixel). Besides reporting PIW and $t$-test results similar to Table \ref{tab:cls_piw_ttest}, we also compute PAvPU, and compare the results with baseline models in \citet{cdropout}, \textit{i.e.}, MC Dropout \citep{mcdropout}, Gaussian Dropout \citep{gausdropout}, Concrete Dropout \citep{concretedropout}, Bayes by Backprop \citep{bbb}, and variants of Contextual Dropout \citep{cdropout}. 

Following the same experimental settings in \citet{cdropout}, we apply a batch size of $128$ for model training, and adopt the same MLP architecture with two hidden layers of $300$ and $100$ hidden units, respectively, each followed by a ReLU non-linearity, as the network architecture for the pre-trained classifier $f_{\phi}(\vx)$. We simplify the diffusion network architecture (detailed in Section~\ref{ssec:general_experiment_details}) by reducing feature dimension from $2048$ to $128$ to match the model parameter scale for a fair comparison. We pre-train $f_{\phi}(\vx)$ for $100$ epochs, and obtain a test set accuracy of $85.50\%$. We train the diffusion model for $1000$ epochs.  Unlike on CIFAR-10 or FashionMNIST dataset, we do not apply data normalization on the noisy MNIST dataset, following the settings in \citet{cdropout}.

Similar to Table \ref{tab:cls_piw_ttest}, we report the mean PIW among correct and incorrect predictions, and the mean accuracy among instances rejected and not-rejected by the paired two-sample $t$-test, in~Table~\ref{tab:cls_noisy_mnist_piw_ttest}. Same as \citet{cdropout}, we set $\alpha=0.05$. We report these metrics at the scope of the entire test set, and for each true class label along with their group accuracy.

\begin{table}[ht]
\caption{\label{tab:cls_noisy_mnist_piw_ttest}PIW (multiplied by $100$) and $t$-test ($\alpha=0.05$) results for Noisy MNIST classification task.}\vspace{1mm}
\begin{center}
\resizebox{9cm}{!}{
\begin{tabular}{@{}l|c|cc|cc@{}}
\toprule[1.5pt]
Class & Accuracy  &       \multicolumn{2}{c|}{PIW}           &   \multicolumn{2}{c}{Accuracy by $t$-test Status}                \\
      &           & Correct  & Incorrect & Rejected  & Not-Rejected (Count)   \\ \midrule
All   & $88.26\%$ & $39.76$ & $76.25$  & $91.24\%$ & $43.04\%$ (618) \\ \midrule
1     & $94.69\%$ & $15.93$ & $68.67$  & $96.63\%$ & $35.48\%$ (31)   \\
2     & $96.48\%$ & $20.48$ & $74.73$  & $97.66\%$ & $46.15\%$ (26)    \\
3     & $88.47\%$ & $29.92$ & $74.98$  & $91.25\%$ & $44.26\%$ (61) \\
4     & $87.23\%$ & $41.38$ & $78.58$  & $90.85\%$ & $45.68\%$ (81)  \\
5     & $85.34\%$ & $48.87$ & $78.61$  & $87.65\%$ & $53.73\%$ (67) \\
6     & $85.76\%$ & $51.35$ & $76.43$  & $89.85\%$ & $40.54\%$ (74)         \\
7     & $92.28\%$ & $22.92$ & $74.04$  & $94.49\%$ & $28.13\%$ (32) \\
8     & $86.97\%$ & $39.80$ & $74.42$  & $90.17\%$ & $37.10\%$ (62)   \\
9     & $80.90\%$ & $62.48$ & $77.73$  & $84.25\%$ & $45.88\%$ (85)         \\
10    & $83.25\%$ & $76.92$ & $76.67$  & $87.69\%$ & $42.42\%$ (99)     \\ \bottomrule[1.5pt]
\end{tabular}
}
\end{center}
\end{table}

From Table \ref{tab:cls_noisy_mnist_piw_ttest}, we are able to draw similar conclusions as in Table \ref{tab:cls_piw_ttest}: across the entire test set, mean PIW of the true class label among the correct predictions is much narrower than that of the incorrect predictions, implying that CARD is confident in its correct predictions, and tends to make mistakes when being vague. When comparing mean PIWs at true class level, we observe that a more accurate class is inclined to have a larger difference between correct and incorrect predictions. Meanwhile, similar to Table \ref{tab:cls_piw_ttest}, the accuracy of test instances rejected by the $t$-test is much higher than that of the not-rejected ones, both across the entire test set and at the class level, while there are almost $10$ times of not-rejected cases as those for CIFAR-10 task. This result could have a more significant practical impact when applying human-machine collaboration: we are able to identify more than $6\%$ of the data with a mean accuracy of less than $50\%$ --- if we pass these cases to human agents, we would be able to remarkably improve the classification accuracy, while still enjoying the automation by the machine classifier in the vast majority of test instances.

Furthermore, we contextualize the performance of CARD by reporting accuracy, PAvPU and NLL, along with those from the baseline models mentioned at the beginning of Section \ref{ssec:noisy_mnist}, in Table \ref{tab:cls_noisy_mnist_pavpu}. The metrics of the other models are from Table 1 in \citet{cdropout}.

\begin{table}[ht]
\caption{\label{tab:cls_noisy_mnist_pavpu}PAvPU (in $\%$) along with accuracy (in $\%$) and NLL for Noisy MNIST classification.}\vspace{1mm}

\begin{center}
\resizebox{8cm}{!}{
\begin{tabular}{@{}l|ccc@{}}
\toprule[1.5pt]
Method                      & Accuracy & PAvPU $(\alpha=0.05)$ & NLL    \\ \midrule
MC - Bernoulli              & $86.36$  & $85.63$        & $1.72$ \\
MC - Gaussian               & $86.31$  & $85.64$        & $1.72$ \\
Concrete                    & $86.52$  & $86.77$        & $1.68$ \\
Bayes by Backprop           & $86.55$  & $87.13$        & $2.30$ \\
Contextual Gating           & $86.20$  & -              & $1.81$ \\
Contextual Gating + Dropout & $86.70$  & $87.01$        & $1.71$ \\
Bernoulli Contextual        & $87.43$  & $87.81$        & $1.41$ \\
Gaussian Contextual         & $87.35$  & $87.72$        & $1.43$ \\ \midrule
CARD (ours)                 & $\bm{88.26}$        & $\bm{89.12}$              & $\bm{0.39}$      \\ \bottomrule[1.5pt]
\end{tabular}
}
\end{center}
\end{table}

We observe that CARD obtains an accuracy of $88.26\%$ (improves from $85.50\%$ by the pre-trained classifier $f_{\phi}$), and a PAvPU of $89.12\%$, both are the best among all models. This implies that our model is making not only accurate classifications, but also high-quality predictions in terms of model confidence. Lastly, we apply temperature scaling \citep{calibration} to calibrate the predicted probability for the computation of NLL, where the temperature parameter is tuned with the training set, and again obtain the best metric among all models.

\subsection{Classification on Large-Scale Benchmark Datasets}\label{ssec:cifar100_cls}
In this section, we demonstrate the generalizability of CARD on classification tasks with our proposed framework of evaluating instance-level model prediction confidence, particularly on large-scale datasets, by reporting the experimental results on CIFAR-100 ($100$ classes, $10,000$ test instances), ImageNet-100 ($100$ classes, $5,000$ test instances), and ImageNet ($1000$ classes, $50,000$ test instances).

For the CIFAR-100 dataset, we pre-train the deterministic base classifier $f_\phi$ with a ResNet-18 architecture, which achieves a test accuracy of $71.37\%$. For the ImageNet-100 dataset, we apply a ResNet-50 architecture and use the parameters of a linear classifier fine-tuned with the self-supervised recipe \citep{he2020momentum,zheng2021contrastive} for $f_\phi$, which achieves a test accuracy of $82.30\%$. For the ImageNet dataset, we adopt three different training paradigms for $f_\phi$, all using a ResNet-50 architecture, but obtain different test accuracies: the first one has its parameters learned with the self-supervised pipeline, \textit{i.e.}, the feature encoder of the ResNet-50 network is firstly pre-trained in a self-supervised manner with the loss proposed by \citet{zheng2021contrastive}, and then by fixing the encoder the linear classifier of the network is further tuned with label supervision --- this $f_\phi$ achieves a test accuracy of $73.87\%$; the second one reads the pre-trained weights provided by TorchVision, with the paradigm specified by \citet{imagenetresnet50_1}, which achieves a test accuracy of $76.13\%$; the third one also reads the pre-trained weights provided by TorchVision, with the paradigm specified by \citet{imagenetresnet50_2}, which achieves a test accuracy of $80.30\%$.

For each of the three large-scale benchmark datasets, we conduct $10$ experimental runs, and report both majority-voted accuracy and PAvPU (with $\alpha=0.01$ for the paired two-sample $t$-test) by CARD, along with the accuracy by the corresponding $f_\phi$, in Table \ref{tab:clf_metrics_largescale_datasets_multiple_runs}. We observe that under all circumstances, CARD is able to improve the accuracy from the base classifier $f_{\phi}$.

For the CIFAR-100 dataset with CARD, we also contextualize the performance of CARD through the accuracy of other BNNs with ResNet-18 architecture in Table~\ref{tab:cifar100_cls_acc_comparison}, where the metrics were reported in \citet{newelbo}.

\begin{table}[ht]
\caption{\label{tab:cifar100_cls_acc_comparison}Comparison of accuracy (in $\%$) on CIFAR-100 dataset with other BNNs.}\vspace{-3mm}
\begin{center}
\resizebox{\textwidth}{!}{
\begin{tabular}{@{}l|cccccccc@{}}
\toprule[1.5pt]
Model         & CMV-MF-VI & CM-MF-VI  & CV-MF-VI  & MF-VI     & MC Dropout & MAP & $f_{\phi}$ (ResNet-18) & CARD \\ \midrule
Accuracy & $60.59\pm 0.39$ & $59.61\pm 0.37$ & $46.22\pm 0.54$ & $40.54\pm 0.72$    & $54.49\pm 0.36$ & $52.08\pm 0.34$ & $71.37$ & $\bm{71.42\pm 0.01}$ \\ \bottomrule[1.5pt]
\end{tabular}}
\end{center}
\end{table}

Regarding instance-level model prediction uncertainty assessment, we pick one run for each dataset, and report PIW among correct and incorrect predictions, at the entire test set level and at the true class level. Since each dataset has either $100$ or $1000$ total number of classes, we only report PIW for the class with the most and the least accurate predictions, as well as the mean accuracy given the $t$-test rejection status at the whole test set level. Furthermore, we record the test accuracy when making the prediction for each instance merely by the class with the narrowest PIW. We report these metrics for all three large-scale datasets in Table \ref{tab:large_scale_datasets_cls_piw_ttest}. For all experimental runs, we are able to draw conclusions consistent with those from the smaller-scale benchmark datasets (CIFAR-10 with Table \ref{tab:cls_piw_ttest}, FashionMNIST with Table \ref{tab:fmnist_cls_piw_ttest}, and Noisy MNIST with Table \ref{tab:cls_noisy_mnist_piw_ttest}): relative variability in label reconstruction captured by true label PIW is strongly related to prediction correctness, with sharper contrast between correct and incorrect predictions in a more accurate class; meanwhile, the accuracy of instances not rejected by the $t$-test is much lower than that of the rejected ones, indicating a great potential of improvement in prediction accuracy if combined with human inspection for the not-rejected cases.

\begin{table}[ht]
\caption{\label{tab:clf_metrics_largescale_datasets_multiple_runs}Majority-voted accuracy and PAvPU by CARD on CIFAR-100, ImageNet-100, and ImageNet, over $10$ experimental runs. Pre-trained base classifier $f_{\phi}$ accuracy is also reported.}\vspace{1mm}
\begin{center}
\resizebox{7.5cm}{!}{
\begin{tabular}{@{}c|c|c|c@{}}
\toprule
Dataset & Acc. by $f_{\phi}$ & Accuracy & PAvPU \\ \midrule\midrule
CIFAR-100     & $71.37\%$ & $71.42\pm 0.01\%$ & $71.48\pm 0.03\%$ \\ \midrule
ImageNet-100     & $82.30\%$ & $82.35\pm 0.03\%$ & $82.73\pm 0.07\%$ \\ \midrule
ImageNet     & $73.87\%$ & $74.28\pm 0.01\%$ & $74.63\pm 0.02\%$ \\ \midrule
ImageNet     & $76.13\%$ & $76.20\pm 0.00\%$ & $76.29\pm 0.01\%$ \\ \midrule
ImageNet     & $80.30\%$ & $80.35\pm 0.01\%$ & $80.55\pm 0.01\%$ \\ 
\bottomrule[1.5pt]
\end{tabular}
}
\end{center}
\end{table}

\begin{table}[ht]
\caption{\label{tab:large_scale_datasets_cls_piw_ttest}PIW (multiplied by $100$), accuracy by predicting with the narrowest PIW, and accuracy by $t$-test rejection status, for the CIFAR-100, ImageNet-100, and ImageNet classification tasks, over a single experimental run. Due to the number of classes, we only report the PIW for all test instances, and within the most and least accurate classes. We applied multiple $f_{\phi}$ for the ImageNet dataset.}\vspace{1mm}
\begin{center}
\resizebox{11cm}{!}{
\begin{tabular}{@{}c|c|cc|c|cc@{}}
\toprule
Dataset & Accuracy  &       \multicolumn{2}{c|}{PIW} & Acc. by PIW & \multicolumn{2}{c}{Acc. by $t$-test Result} \\ 
      &           & Correct  & Incorrect & & Rejected & Not-Rejected (Count) \\ \midrule\midrule
\multicolumn{7}{l}{CIFAR-100} \\ \midrule
overall   & $71.42\%$ & $0.59$ & $3.91$ & $60.53\%$ & $71.56\%$ & $35.90\%$ (39) \\ \midrule 
most acc.     & $95.00\%$ & $0.16$ & $1.92$ \\
least acc.     & $44.00\%$ & $5.09$ & $5.84$ \\ \midrule\midrule
\multicolumn{7}{l}{ImageNet-100} \\ \midrule
overall   & $82.34\%$ & $2.06$ & $13.73$ & $68.64\%$ & $82.90\%$ & $34.48\%$ (58) \\ \midrule 
most acc.     & $98.00\%$ & $0.72$ & $8.06$ \\
least acc.     & $42.00\%$ & $6.79$ & $14.15$ \\ \midrule\midrule
\multicolumn{7}{l}{ImageNet ($f_{\phi}$ Accuracy $73.87\%$)} \\ \midrule
overall   & $74.28\%$ & $0.65$ & $3.11$ & $69.22\%$ & $74.63\%$ & $24.93\%$ (349) \\ \midrule 
most acc.     & $98.00\%$ & $0.27$ & $2.80$ \\
least acc.     & $8.00\%$ & $20.10$ & $50.07$ \\ \midrule\midrule
\multicolumn{7}{l}{ImageNet ($f_{\phi}$ Accuracy $76.13\%$)} \\ \midrule
overall   & $76.20\%$ & $0.51$ & $3.60$ & $75.21\%$ & $76.30\%$ & $25.71\%$ (105) \\ \midrule 
most acc.     & $98.00\%$ & $0.08$ & $2.66$ \\
least acc.     & $18.00\%$ & $1.87$ & $3.26$ \\ \midrule\midrule
\multicolumn{7}{l}{ImageNet ($f_{\phi}$ Accuracy $80.30\%$)} \\ \midrule
overall   & $80.35\%$ & $1.42$ & $5.13$ & $74.08\%$ & $80.59\%$ & $27.63\%$ (228) \\ \midrule 
most acc.     & $98.00\%$ & $0.49$ & $2.34$ \\
least acc.     & $8.00\%$ & $91.70$ & $84.61$ \\ \midrule 
\bottomrule[1.5pt]
\end{tabular}
}
\end{center}
\end{table}

\subsection{General Experiment Setup Details}\label{ssec:general_experiment_details}
In this section, we provide the experimental setup for the CARD model in both regression and classification tasks.

\textbf{Training}: As mentioned at the beginning of Section \ref{sec:card_experiments}, we set the number of timesteps to $1000$, and adopt a linear $\beta_t$ schedule same as \citet{ddpm}. We set the learning rate to $0.001$ for all tasks. We use the AMSGrad \citep{amsgrad} variant of the Adam optimizer \citep{adam} for all regression tasks. We use the Adam optimizer for the classification tasks on all presented datasets, and adopt cosine learning rate decay \citep{cosinedecay}. We use exponentially weighted moving averages \citep{ewma} on model parameters with a decay factor of $0.9999$. We adopt antithetic sampling \citep{antitheticsampling} to draw correlated timesteps during training. We set the number of epochs at $5000$ by default for regression tasks, and $1000$ by default for classification tasks, to sufficiently cover the timesteps with $T=1000$. The batch size for each UCI regression task are reported in Table \ref{tab:uci_batch_size_settings}. We also apply a batch size of $256$ for all toy regression tasks and classification tasks. For all UCI regression tasks, we follow the convention in \citet{pbp} and standardize both the input features and the response variable to have zero mean and unit variance, and remove the standardization to compute the metrics. For all toy regression tasks, we do not standardize the input feature; we only standardize the response variable on the log-log cubic regression task. For classification on CIFAR-10 and FashionMNIST, we normalize the dataset with the mean and standard deviation of the training set.

\textbf{Network Architecture}: For the diffusion model, we adopt a simpler network architecture than that in previous work \citep{xiao2021tackling, truncdiff}, by first changing the Transformer sinusoidal position embedding to linear embedding for the timestep. As the network $\epsilonv_{\rvtheta}\big(\vx, \vy_t, f_{\phi}(\vx), t\big)$ has three other inputs besides the timestep $t$, we integrate them in different ways for regression and classification~tasks: 

\begin{itemize}
    \item For regression, we first concatenate $\vx$, $\vy_t$, and $f_{\phi}(\vx)$, then send the resulting vector through three fully-connected layers, all with an output dimension of $128$. We perform Hadamard product between each of the output vector with the corresponding timestep embedding, followed by a Softplus non-linearity, before sending the resulting vector to the next fully-connected layer. Lastly, we apply a $4$-th fully-connected layer to map the vector to one with a dimension of $1$, as the output forward diffusion noise prediction. We summarize the architecture in Table \ref{tab:network_arch_regression} (a).
    \item For classification on CIFAR-10, we first apply an encoder on the flattened input image (originally $32\times32\times3$) to obtain a representation with $4096$ dimensions. The encoder consists of three fully-connected layers with an output dimension of $4096$. Meanwhile, we concatenate $\vy_t$ and $f_{\phi}(\vx)$, and apply a fully-connected layer to obtain an output vector of $4096$ dimensions. We perform a Hadamard product between such vector and a timestep embedding to obtain a response embedding conditioned on the timestep. We then perform Hadamard product between image embedding and response embedding to integrate these variables, and send the resulting vector through two more fully-connected layers with $4096$ output dimensions, each would first followed by a Hadamard product with a timestep embedding, and lastly a fully-connected layer with an output dimension of $1$ as the noise prediction. Note that all fully-connected layers are also followed by a batch normalization layer and a Softplus non-linearity, except the output layer. We summarize the architecture in Table \ref{tab:network_arch_regression} (b).
\end{itemize}

\begin{table}[!ht]
\centering
\caption{CARD $\epsilonv_{\rvtheta}$ network architecture. We denote concatenation as $\oplus$, Hadamard product as $\odot$, and Softplus non-linearity as $\sigma$. We also denote a fully-connected layer as $g$, and a hidden layer output as $l$, with subscripts to differentiate them.}\vspace{1mm}
\label{tab:network_arch_regression}
\begin{minipage}[t]{0.5\linewidth}\centering
{(a) Regression network architecture.}
\addstackgap[5pt]{
\resizebox{5.5cm}{!}{
\begin{tabular}{c}
\toprule[1.5pt] 
input: $\vx, \vy_t, f_{\phi}(\vx), t$ \\ \cmidrule(lr){1-1}
$l_1=\sigma\Big(g_{1,a}\big(\vx\oplus\vy_t\oplus f_{\phi}(\vx)\big)\odot g_{1,b}(t)\Big)$ \\ \cmidrule(lr){1-1}
$l_2=\sigma\big(g_{2,a}(l_1)\odot g_{2,b}(t)\big)$ \\ \cmidrule(lr){1-1}
$l_3=\sigma\big(g_{3,a}(l_2)\odot g_{3,b}(t)\big)$ \\ \cmidrule(lr){1-1}
output: $g_{4}(l_3)$ \\
\bottomrule[1.5pt]
\end{tabular}
}
}
\end{minipage}\hfill%
\begin{minipage}[t]{0.5\linewidth}\centering
{(b) Classification network architecture.}
\addstackgap[5pt]{
\resizebox{6cm}{!}{
\begin{tabular}{c}
\toprule[1.5pt] 
input: $\vx, \vy_t, f_{\phi}(\vx), t$ \\ \cmidrule(lr){1-1}
$l_{1,x}=\sigma\Big(\text{BN}\big(g_{1, x}(\vx)\big)\Big)$ \\ \cmidrule(lr){1-1}
$l_{2,x}=\sigma\Big(\text{BN}\big(g_{2, x}(\vx)\big)\Big)$ \\ \cmidrule(lr){1-1}
$l_{3,x}=\text{BN}\big(g_{1, x}(\vx)\big)$ \\ \cmidrule(lr){1-1}
$l_{1,y}=\sigma\bigg(\text{BN}\Big(g_{1,y}\big(\vy_t\oplus f_{\phi}(\vx)\big)\odot g_{1,b}(t)\Big)\bigg)$ \\ \cmidrule(lr){1-1}
$l_1=l_{3,x}\odot l_{1,y}$ \\ \cmidrule(lr){1-1}
$l_2=\sigma\Big(\text{BN}\big(g_{2,a}(l_1)\odot g_{2,b}(t)\big)\Big)$ \\ \cmidrule(lr){1-1}
$l_3=\sigma\Big(\text{BN}\big(g_{3,a}(l_2)\odot g_{3,b}(t)\big)\Big)$ \\ \cmidrule(lr){1-1}
output: $g_{4}(l_3)$ \\
\bottomrule[1.5pt]
\end{tabular}
}
}
\end{minipage}
\end{table}

For the pre-trained model $f_{\phi}(\vx)$, we adjust the functional form and training scheme based on the task. For regression, we adopt a feed-forward neural network with two hidden layers, each with $100$ and $50$ hidden units, respectively. We apply a Leaky ReLU non-linearity with a $0.01$ negative slope after each hidden layer. In practice, we find that when the dataset size is small (most of the UCI tasks have less than $10,000$ data points, and several around or less than $1000$), a deep neural network $f_{\phi}(\vx)$ is prone to overfitting. We thus set the default number of epochs to $1000$, and adopt early stopping with a patience of $50$ epochs, \textit{i.e.}, we terminate the training if there is no improvement in the validation set MSE for $50$ consecutive epochs. We split the original training set with $60\%/40\%$ ratio and apply early stopping to find the optimal number of epochs, then train $f_{\phi}(\vx)$ on the full training set. For classification, we apply a pre-trained ResNet-18 network for CIFAR-10 dataset. We only train the model with $10$ epochs, as more would lead to overfitting. We apply the Adam optimizer for $f_{\phi}(\vx)$ in all tasks, and we only pre-train the model and freeze it during the training of the diffusion model, instead of fine-tuning it.

\subsection{UCI Baseline Model Experiment Setup Details and Dataset Information}\label{ssec:uci_baseline_experiment_details}
In this section, we first provide the experimental setup for the UCI regression baseline models (PBP, MC Dropout, Deep Ensembles, and GCDS), including learning rate, batch size, network architecture, number of epochs, etc. We applied the GitHub repo \citep{nnuncert} to run BNN models, and implemented our own version of GCDS \citep{jointmatching} since the code has not been published. We note that GCDS is related to a concurrent work of \citet{yang2022regularized}, who share a comparable idea but use it in a different application: regularizing the learning of an implicit policy in offline reinforcement~learning.

Overall, we apply the same learning rate of $0.001$ for all models on all datasets, except for PBP on the Boston dataset, which is $0.1$. We also apply the Adam optimizer for all experiments. We follow the convention in \citet{pbp} and standardize both the input features and response variable for training, and remove the standardization for evaluation.

For batch size, we adjust on a case-by-case basis, taking the running time and dataset size into consideration. We provide the batch size in Table \ref{tab:uci_batch_size_settings}. For network architecture, we use ReLU non-linearities for all $3$ BNNs, and Leaky ReLU with a $0.01$ negative slope for GCDS; we choose the number of hidden layers with the number of hidden units per layer from the following $3$ options: a) $1$ hidden layer with $50$ hidden units; b) $1$ hidden layer with $100$ hidden units; c) $2$ hidden layers with $100$ and $50$ hidden units, respectively. We show the choice of hidden layer architecture in Table \ref{tab:uci_net_arch_settings}. For the number of training epochs, we also vary on a case-by-case basis: PBP and Deep Ensembles both applied $40$ epochs, but we observed in many experiments that the model has not converged. We show the number of epochs in Table \ref{tab:uci_n_epochs_settings}. Lastly, datasets Boston, Energy, and Naval all contain one or more categorical variables, thus we ran the experiments both with and without conducting one-hot encoding on the data. We found that except the Naval dataset with PBP, all other cases had worse metrics when one-hot encoding was applied.

We summarize the dataset information in terms of their size and number of features in Table~\ref{tab:reg_uci_dim}.

\begin{table}[ht]
\caption{\label{tab:reg_uci_dim}Dataset size ($N$ observations, $P$ features) of UCI regression tasks.}\vspace{-2.5mm}
\begin{center}
\resizebox{\textwidth}{!}{
\begin{tabular}{@{}c|cccccccccc@{}}
\toprule[1.5pt]
Dataset  & Boston      & Concrete     & Energy    & Kin8nm       & Naval          & Power        & Protein       & Wine          & Yacht     & Year            \\ \midrule
$(N, P)$ & $(506, 13)$ & $(1030,8)$ & $(768,8)$ & $(8192,8)$ & $(11,934,16)$ & $(9568,4)$ & $(45,730,9)$ & $(1599,11)$ & $(308,6)$ & $(515,345,90)$     \\ \bottomrule[1.5pt]
\end{tabular}}
\end{center}
\end{table}

\begin{table}[ht]
\caption{\label{tab:uci_batch_size_settings}Batch size settings of UCI regression tasks across different models.}\vspace{1mm}
\begin{center}
\resizebox{9cm}{!}{
\begin{tabular}{@{}l|ccccc@{}}
\toprule[1.5pt]
         & PBP   & MC Dropout & Deep Ensembles & GCDS  & CARD (ours) \\ \midrule
Boston   & $32$  & $32$       & $32$           & $32$  & $32$        \\
Concrete & $32$  & $32$       & $32$           & $32$  & $32$        \\
Energy   & $32$  & $32$       & $32$           & $32$  & $32$        \\
Kin8nm   & $64$  & $32$       & $64$           & $64$  & $64$        \\
Naval    & $64$  & $32$       & $64$           & $64$  & $64$        \\
Power    & $64$  & $64$       & $64$           & $64$  & $64$        \\
Protein  & $100$ & $256$      & $100$          & $256$ & $256$       \\
Wine     & $32$  & $32$       & $32$           & $32$  & $32$        \\
Yacht    & $32$  & $32$       & $32$           & $32$  & $32$        \\
Year     & $256$ & $256$      & $100$          & $256$ & $256$  \\ \bottomrule[1.5pt]
\end{tabular}
}
\end{center}
\end{table}

\begin{table}[ht]
\caption{\label{tab:uci_net_arch_settings}Network hidden layer architecture for UCI regression tasks across different models.}\vspace{1mm}
\begin{center}
\resizebox{7.5cm}{!}{
\begin{tabular}{@{}l|cccc@{}}
\toprule[1.5pt]
         & \multicolumn{1}{l}{PBP} & \multicolumn{1}{l}{MC Dropout} & \multicolumn{1}{l}{Deep Ensembles} & \multicolumn{1}{l}{GCDS} \\ \midrule
Boston   & a                       & c                              & a                                  & c                        \\
Concrete & a                       & c                              & c                                  & c                        \\
Energy   & a                       & b                              & a                                  & c                        \\
Kin8nm   & a                       & c                              & a                                  & a                        \\
Naval    & a                       & c                              & a                                  & a                        \\
Power    & a                       & b                              & a                                  & a                        \\
Protein  & b                       & c                              & b                                  & c                        \\
Wine     & a                       & c                              & a                                  & c                        \\
Yacht    & a                       & a                              & a                                  & a                        \\
Year     & b                       & c                              & b                                  & b                        \\ \bottomrule[1.5pt]
\end{tabular}
}
\end{center}
\end{table}

\begin{table}[ht]
\caption{\label{tab:uci_n_epochs_settings}Number of training epochs of UCI regression tasks across different models.}\vspace{1mm}
\begin{center}
\resizebox{7.5cm}{!}{
\begin{tabular}{@{}l|cccc@{}}
\toprule[1.5pt]
         & \multicolumn{1}{l}{PBP} & \multicolumn{1}{l}{MC Dropout} & \multicolumn{1}{l}{Deep Ensembles} & \multicolumn{1}{l}{GCDS} \\ \midrule
Boston   & $100$                   & $1000$                         & $100$                              & $500$                    \\
Concrete & $100$                   & $500$                          & $40$                               & $500$                    \\
Energy   & $100$                   & $500$                          & $100$                              & $500$                    \\
Kin8nm   & $100$                   & $500$                          & $100$                              & $500$                    \\
Naval    & $100$                   & $500$                          & $100$                              & $500$                    \\
Power    & $100$                   & $500$                          & $100$                              & $500$                    \\
Protein  & $100$                   & $500$                          & $100$                              & $500$                    \\
Wine     & $100$                   & $500$                          & $100$                              & $500$                    \\
Yacht    & $100$                   & $500$                          & $100$                              & $500$                    \\
Year     & $100$                   & $100$                          & $100$                              & $500$                    \\ \bottomrule[1.5pt]
\end{tabular}
}
\end{center}
\end{table}

We reiterate that we re-ran the experiments with the baseline BNN models to compute the new metric QICE along with the conventional metrics RMSE and NLL. We carefully tuned the hyperparameters to obtain results better than or comparable with the ones reported in the original papers.

\subsection{UCI Regression Tasks PICP across All Methods}
In this section, we report PICP for all methods in Table \ref{tab:reg_uci_picp} from the same runs with the corresponding metrics in Tables \ref{tab:reg_uci_rmse}, \ref{tab:reg_uci_nll}, and \ref{tab:reg_uci_qice}.

\begin{table}[ht]
\caption{\label{tab:reg_uci_picp}PICP (in $\%$) of UCI regression tasks.}\vspace{1mm}
\begin{center}
\resizebox{10cm}{!}{
\begin{tabular}{@{}l|ccccc@{}}
\toprule[1.5pt]
Dataset             &                    &                    & $\mid\text{PICP} - 95\mid$ $\downarrow$               &                    &                    \\
                    & PBP                & MC Dropout         & Deep Ensembles     & GCDS               & CARD (ours)               \\ \midrule
Boston      & $91.27\pm 4.82$ & $\bm{96.08\pm 2.70}$ & $88.73\pm 5.68$ & $31.37\pm 6.79$ & $93.24\pm 3.59$ \\
Concrete            & $92.28\pm 2.87$ & $\bm{97.52\pm 2.43}$ & $90.34\pm 3.64$ & $39.85\pm 4.53$ & $90.24\pm 3.45$ \\
Energy              & $93.18\pm 3.12$ & $99.03\pm 1.08$ & $\bm{96.49\pm 1.97}$ & $63.57\pm 10.26$ & $98.70\pm 1.30$ \\
Kin8nm$^1$              & $\bm{95.06\pm 0.77}$ & $95.37\pm 2.24$ & $96.53\pm 0.67$ & $59.06\pm 5.31$ & $93.68\pm 0.79$ \\
Naval$^2$    & $93.52\pm 4.40$ & $100.00\pm 0.00$ & $99.78\pm 0.28$ & $83.71\pm 14.87$ & $\bm{95.35\pm0.60}$  \\
Power         & $95.75\pm 0.69$ & $96.28\pm 0.76$ & $95.91\pm 0.71$ & $89.13\pm 1.14$ & $\bm{94.87\pm 0.65}$ \\
Protein             & $\bm{94.79\pm 0.13}$ & $96.46\pm 0.77$ & $96.08\pm 0.28$ & $85.24\pm 0.86$ & $95.38\pm 0.16$ \\
Wine                & $92.72\pm 1.80$ & $91.41\pm 2.66$ & $91.06\pm 1.64$ & $86.37\pm 2.33$ & $\bm{93.88\pm 2.10}$ \\
Yacht               & $\bm{96.94\pm 2.60}$ & $100.00\pm 0.00$ & $98.87\pm 1.54$ & $83.23\pm 4.28$ & $99.84\pm 0.70$ \\
Year & $93.04\pm$ NA      & $\bm{94.61\pm}$ NA            & $95.44\pm$ NA      & $87.08\pm$ NA      & $93.35\pm$ NA      \\ \midrule
\# best & $\bm{3}$      & $\bm{3}$            & $1$      & $0$      & $\bm{3}$      \\ \bottomrule[1.5pt]
\end{tabular}
}
\end{center}
\end{table}

\subsection{Ablation Study on Choice of Prior --- UCI Boston Dataset}
In this section, we conduct ablation study for two model variants with different prior distribution settings on the UCI Boston dataset, under various settings of the number of timesteps $T$ with adjusted $\beta_t$ linear schedule (to make sure that $\sqrt{\bar{\alpha}_1}$ is close to $1$ and $\sqrt{\bar{\alpha}_T}$ is close to $0$). We report the evaluation metrics in Table \ref{tab:boston_ablation}, where we compare the original CARD setting with $\mathcal{N}(f_{\phi}(\vx), \mI)$ as the prior distribution at timestep $T$, to the alternative setting with $\mathcal{N}(\mathbf{0}, \mI)$ as the prior distribution. We observe that as $T$ decreases, RMSE and NLL do not deteriorate for $\mathcal{N}(f_{\phi}(\vx), \mI)$ prior (CARD setting), but those from $\mathcal{N}(\mathbf{0}, \mI)$ prior become worse. The metrics that measure distributional fitting, QICE and PICP, gets worse under the $\mathcal{N}(f_{\phi}(\vx), \mI)$ prior setting as well, but such deterioration is not as much as $\mathcal{N}(\mathbf{0}, \mI)$ prior. The results indicate that our setting of an informative prior $\mathcal{N}(f_{\phi}(\vx), \mI)$ contributes to the regression performance of CARD. Furthermore, the setting of the total number of timesteps $T$ does not affect the mean estimation for $\mathcal{N}(\mathbf{0}, \mI)$ prior, but would noticeably impact the distributional fitting (\textit{i.e.}, the recovery of aleatoric uncertainty).

\begin{table}[ht]
\caption{\label{tab:boston_ablation}Ablation study on $2$ prior distribution settings on UCI Boston dataset with different $T$.}\vspace{1mm}
\begin{center}
\resizebox{11cm}{!}{
\begin{tabular}{@{}c|c|c|c|c|c|c@{}}
\toprule
T    & \begin{tabular}[c]{@{}c@{}}$\beta_t$ schedule\\ $(\beta_1, \beta_T)$\end{tabular} & Prior & RMSE           & NLL            & QICE           & PICP            \\ \midrule
\multirow{2}{*}{1000} & $\multirow{2}{*}{(0.0001, 0.02)}$                                                                  & $\mathcal{N}(f_{\phi}(\vx), \mI)$      & $\bm{2.61\pm 0.63}$ & $2.65\pm 0.12$ & $\bm{3.45\pm 0.83}$ & $93.24\pm 3.59$ \\
     &                                                                                   & $\mathcal{N}(\mathbf{0}, \mI)$      & $2.71\pm 0.69$ & $\bm{2.37\pm 0.12}$ & $3.53\pm 0.99$ & $\bm{93.53\pm 3.34}$ \\\midrule
\multirow{2}{*}{500}  & $\multirow{2}{*}{(0.0001, 0.04)}$                                                                  & $\mathcal{N}(f_{\phi}(\vx), \mI)$      & $\bm{2.63\pm 0.72}$ & $\bm{2.33\pm 0.13}$ & $3.94\pm 1.05$ & $\bm{93.14\pm 3.19}$ \\
     &                                                                                   & $\mathcal{N}(\mathbf{0}, \mI)$      & $2.70\pm 0.68$ & $2.34\pm 0.12$ & $\bm{3.48\pm 0.76}$ & $91.76\pm 3.75$ \\\midrule
\multirow{2}{*}{100}  & $\multirow{2}{*}{(0.001, 0.175)}$                                                                  & $\mathcal{N}(f_{\phi}(\vx), \mI)$      & $\bm{2.65\pm 0.67}$ & $\bm{2.30\pm 0.18}$ & $\bm{4.09\pm 1.13}$ & $\bm{88.82\pm 5.15}$ \\
     &                                                                                   & $\mathcal{N}(\mathbf{0}, \mI)$      & $2.69\pm 0.66$ & $2.32\pm 0.21$ & $4.19\pm 1.12$ & $85.20\pm 6.34$ \\\midrule
\multirow{2}{*}{50}   & $\multirow{2}{*}{(0.001, 0.35)}$                                                                   & $\mathcal{N}(f_{\phi}(\vx), \mI)$      & $\bm{2.61\pm 0.71}$ & $\bm{2.31\pm 0.25}$ & $\bm{5.06\pm 1.46}$ & $\bm{81.96\pm 6.31}$ \\
     &                                                                                   & $\mathcal{N}(\mathbf{0}, \mI)$      & $2.76\pm 0.66$ & $2.57\pm 0.39$ & $5.38\pm 1.55$ & $76.18\pm 7.13$ \\\midrule
\multirow{2}{*}{10}   & $\multirow{2}{*}{(0.01, 0.95)}$                                                                    & $\mathcal{N}(f_{\phi}(\vx), \mI)$      & $\bm{2.63\pm 0.58}$ & $\bm{2.56\pm 0.44}$ & $\bm{5.34\pm 1.24}$ & $\bm{77.65\pm 7.00}$ \\
     &                                                                                   & $\mathcal{N}(\mathbf{0}, \mI)$      & $2.80\pm 0.75$ & $2.98\pm 0.85$ & $5.52\pm 1.20$ & $75.39\pm 7.58$ \\ \bottomrule
\end{tabular}
}
\end{center}
\end{table}

\subsection{Ablation Study on Diffusion Network Parameterization --- CIFAR-10 Dataset}
In this section, we study the impact of different $\epsilonv_{\rvtheta}$ network parameterization forms on the CIFAR-10 dataset, through model performance in terms of accuracy and PAvPU, as well as training efficiency at the first $100$ epochs. We compare four model variants, each with a different prior and $\epsilonv_{\rvtheta}$ network parameterization combination, in Table \ref{tab:cifar10_ablation}, by reporting accuracy and PAvPU on the test set over $10$~runs.

\begin{table}[ht]
\caption{\label{tab:cifar10_ablation} $4$ Ablation study on 4 model variants on CIFAR-10 dataset.}\vspace{1mm}
\begin{center}
\resizebox{11cm}{!}{
\begin{tabular}{@{}c|c|c|c|c@{}}
\toprule
Variant & Prior                           & $f_{\phi}(\vx)$ included as $\epsilonv_{\rvtheta}$ input & Accuracy        & PAvPU           \\ \midrule
V1      & $\mathcal{N}(f_{\phi}(\vx), \mI)$ & True                                           & $90.93\pm 0.02$ & $\bm{91.11\pm 0.04}$ \\
V2      & $\mathcal{N}(f_{\phi}(\vx), \mI)$ & False                                          & $\bm{90.94\pm 0.02}$ & $91.08\pm 0.03$ \\
V3      & $\mathcal{N}(\mathbf{0}, \mI)$  & True                                           & $90.88\pm 0.03$ & $91.06\pm 0.03$ \\
V4      & $\mathcal{N}(\mathbf{0}, \mI)$  & False                                          & $90.82\pm 0.02$ & $91.02\pm 0.03$ \\ \bottomrule
\end{tabular}
}
\end{center}
\end{table}

We observe that given the same prior distribution setting, both metrics do not differ much by whether or not we include $f_{\phi}(\vx)$ as the input of the $\epsilonv_{\rvtheta}$ network. Meanwhile, both model variants (V1, V2) with a prior of $\mathcal{N}(f_{\phi}(\vx), \mI)$ outperform the other two variants (V3, V4) of $\mathcal{N}(\mathbf{0}, \mI)$ prior, suggesting the application of an informative prior would benefit the performance. Furthermore, the variant (V4) with $f_{\phi}(\vx)$ as neither the prior mean nor $\epsilonv_{\rvtheta}$ input has the worst performance, indicating the inclusion of a pre-trained classifier can improve the model performance in both accuracy and uncertainty~estimation.

\begin{figure*}[t]
    \centering
    \includegraphics[width=3.2in, height=2.25in]{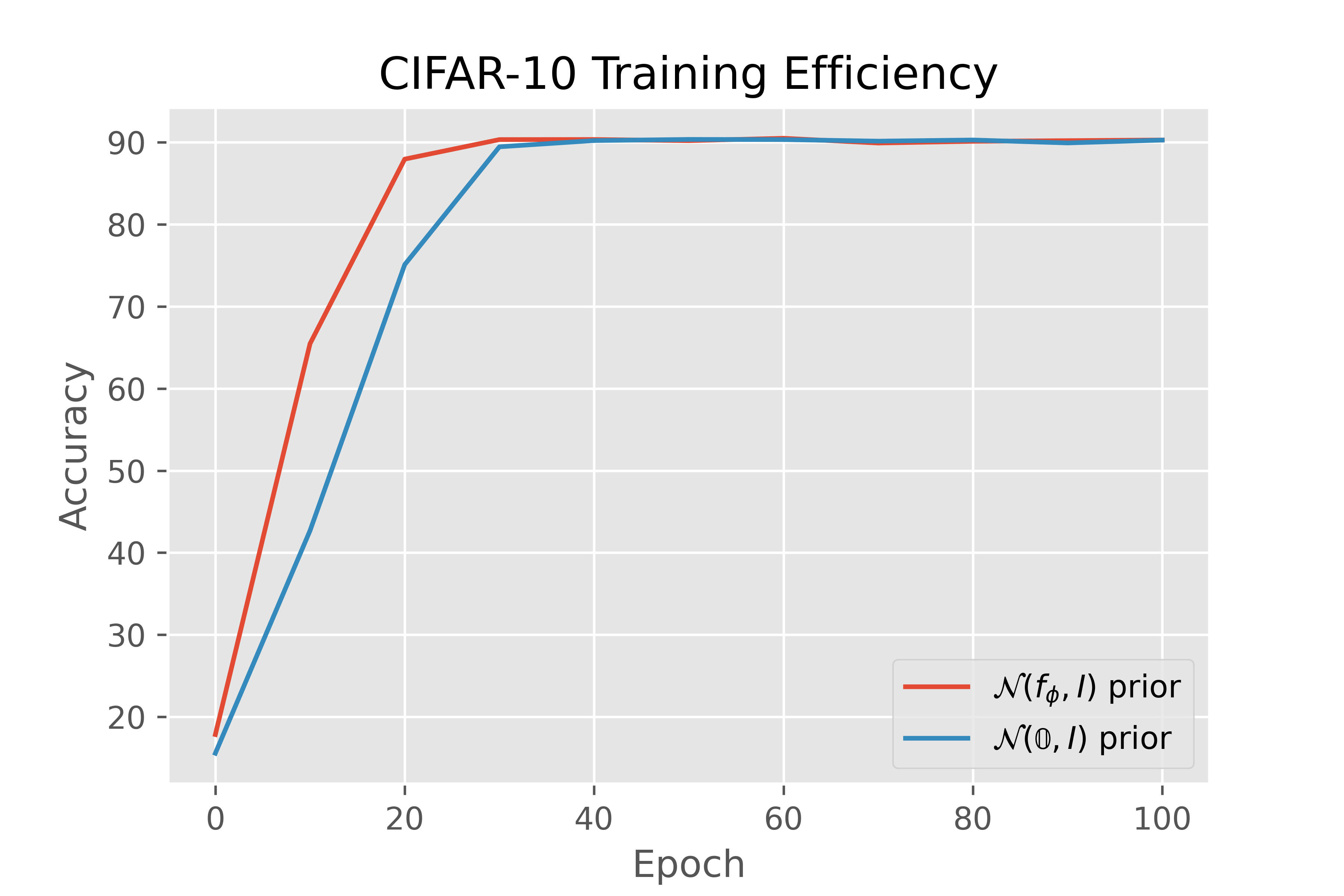}
    \vspace{-2mm}
    \caption{Performance from two prior settings on CIFAR-10 test set with $1$ sample.}
    \label{fig:cifar10_train_efficiency}
\end{figure*}

Furthermore, the choice of $\mathcal{N}(f_{\phi}(\vx), \mI)$ prior also helps with training efficiency: we observe in Figure \ref{fig:cifar10_train_efficiency} that the model performance improved faster for $\mathcal{N}(f_{\phi}(\vx), \mI)$ prior at the beginning of training, by measuring the accuracy on the test set with $1$ sample every $10$ epochs for the first $100$ epochs during training, and plotting the metric (as the mean across all runs) against the number of epochs. Due to the measurement similarity, we omit V2 and V4 and only plot the metrics from V1 (for $\mathcal{N}(f_{\phi}(\vx), \mI)$ prior) and V3 (for $\mathcal{N}(\mathbf{0}, \mI)$ prior). We observe that after only $20$ epochs, the accuracy of $1$ sample by CARD is already close to $90\%$, while the variant with a $\mathcal{N}(\mathbf{0}, \mI)$ prior is only around $75\%$, suggesting the advantage in training efficiency with an informative prior.

\subsection{Regression Toy Example Details}\label{ssec:reg_toy_details}

The $8$ toy examples are summarized by Table \ref{tab:reg_toy_funcs}. For each task, we create the dataset by sampling $10,240$ data points from the data generating function, and randomly split them into training and test sets with an $80\% / 20\%$ ratio. For all uni-modal cases as well as the full circle task, the $\vx$ variable is sampled from a uniform distribution. The noise variable $\epsilonv$ is sampled from a Gaussian distribution. The dataset of the inverse sinusoidal task is created by simply swapping $\vx$ and $\vy$ variable of the sinusoidal task (so that we have multi-modality when the new $\vx$ is roughly between $0.25$ and $0.75$), thus the name of the task.

\begin{table}[ht]
\caption{\label{tab:reg_toy_funcs}Regression toy examples.}\vspace{1mm}
\begin{center}
\resizebox{11cm}{!}{
\begin{tabular}{@{}l|ccc@{}}
\toprule[1.5pt]
Regression Task                     & Data Generating Function                                            & $\vx$        & $\epsilonv$  \\ \midrule
Linear             & $\vy=2\vx+3+\epsilonv$                                                   & $\text{U}(-5, 5)$ & $\mathcal{N}(0, 2^2)$    \\ 
Quadratic          & $\vy=3\vx^2+2\vx+1+\epsilonv$                                              & $\text{U}(-5, 5)$ & $\mathcal{N}(0, 2^2)$    \\ 
Log-Log Linear     & $\vy=\exp\big(\log(\vx)+\epsilonv\big)$                          & $\text{U}(0, 10)$ & $\mathcal{N}(0, 0.15^2)$ \\ 
Log-Log Cubic      & $\vy=\exp\big(3\log(\vx)+\epsilonv\big)$                         & $\text{U}(0, 10)$ & $\mathcal{N}(0, 0.15^2)$ \\ 
Sinusoidal\tablefootnote[1]{The data generating function of this task was originally proposed in \citet{mdn}.}         & $\vy=\vx+0.3\sin(2\pi\vx)+\epsilonv$                                      & $\text{U}(0, 1)$  & $\mathcal{N}(0, 0.08^2)$ \\ 
Inverse Sinusoidal\tablefootnote[2]{Swap $\vx$ and the generated $\vy$ from the sinusoidal regression task.} & swap $\vx$ and $\vy$ from Sinusoidal                                      & \textemdash  & \textemdash \\ 
8 Gaussians\tablefootnote[3]{We set the coordinates of $8$ modes at $(\sqrt{2}, 0)$, $(-\sqrt{2}, 0)$, $(0, \sqrt{2})$, $(0, -\sqrt{2})$, $(1, 1)$, $(1, -1)$, $(-1, 1)$, $(-1, -1)$, and add a noise sample to both coordinates of a mode to generate one instance.}                   & $8$ modes                     & \textemdash          & $\mathcal{N}(0, 0.1^2)$ \\ 
Full Circle                   & $\vy = (10+\epsilonv)\big(\cos(2\pi\vx)+\sin(2\pi\vx)\big)$ & $\text{U}(0, 1)$  & $\mathcal{N}(0, 0.5^2)$  \\ \bottomrule[1.5pt]
\end{tabular}
}
\end{center}
\end{table}

To quantitatively evaluate the performance of CARD, we generate $1000$ $\vy$ samples for each $\vx$ in the test set, and compute the corresponding metrics. We conduct such a procedure over $10$ runs, each applying a different random seed to generate the dataset, and report the mean and standard deviation over all runs for each metric. For all tasks regardless of the form of $p(\vy\given\vx)$, we compute PICP and QICE. For tasks with uni-modal $p(\vy\given\vx)$ distributions, we summarize the $1000$ samples for each test $\vx$ by computing their mean, as an unbiased estimator to $\mathbb{E}(\vy\given\vx)$, and compute the root mean squared error (RMSE) between the estimated and true conditional means. For all tasks, we obtain a mean PICP very close to the optimal $95\%$, and most of the tasks have a mean QICE value far less than $0.01$ except log-log cubic regression, which also has a mean RMSE noticeably larger by an order of magnitude among cases with uni-modal conditional distributions. Note that the $\vy$ samples here have a much wider range: as $\vx$ increases from $0$ to $10$, $\vy$ increases from $0$ to over $1200$, resulting in a much more difficult task. Therefore, the metrics reported here can be viewed with relativity, and combined with the qualitative conclusions from Figure \ref{fig:reg_toy_combined}. The metrics of all tasks, including RMSE, QICE, and PICP, are recorded in Table \ref{tab:reg_toy_metrics}. Note that QICE has been converted to a percentage scale as we report two significant figures for all metrics.

\begin{table}[ht]
\caption{\label{tab:reg_toy_metrics}Regression toy example RMSE, QICE (in $\%$), and PICP (in $\%$).}\vspace{1mm}
\begin{center}
\resizebox{8cm}{!}{
\begin{tabular}{@{}l|ccc@{}}
\toprule[1.5pt]
Regression Task    & RMSE $\downarrow$                       & QICE $\downarrow$              & PICP \\ \midrule
Linear             & $0.07\pm 0.02$         & $0.54\pm 0.14$ & $95.29\pm 0.53$ \\
Quadratic          & $0.21\pm 0.03$         & $0.55\pm 0.12$ & $95.12\pm 0.55$ \\
Log-Log Linear     & $0.07\pm 0.01$         & $0.55\pm 0.15$ & $95.17\pm 0.62$ \\
Log-Log Cubic      & $ 5.85\pm 1.38$        & $1.31\pm 0.26$ & $96.08\pm 0.62$ \\
Sinusoidal         & $0.01\pm 0.00$         & $0.48\pm 0.11$ & $94.81\pm 0.54$ \\
Inverse Sinusoidal & \textemdash & $0.71\pm 0.18$ & $95.89\pm 0.52$ \\
8 Gaussians        & \textemdash & $0.66\pm 0.19$ & $95.92\pm 0.46$ \\
Full Circle        & \textemdash & $0.60\pm 0.05$ & $95.52\pm 0.42$ \\ \bottomrule[1.5pt]
\end{tabular}
}
\end{center}
\end{table}

The results in Table \ref{tab:reg_toy_metrics} implicitly suggest that our proposed metric QICE is reasonable: when RMSE is low (way below $1$) and PICP is close to $95\%$, implying that CARD is performing well in terms of both mean estimation and distributional matching, QICE is also low (far less than $1\%$); for the most difficult task, log-log cubic regression, as RMSE is above $5$ and PICP deviates relatively most from $95\%$ (but not much), QICE also has the largest value (slightly above $1\%$).

\subsection{The Evolution of Samples through the Diffusion Process}
\begin{figure*}[ht]
    \vspace{-3mm}
    \makebox[\textwidth][c]{
      \includegraphics[width=1.2\textwidth]{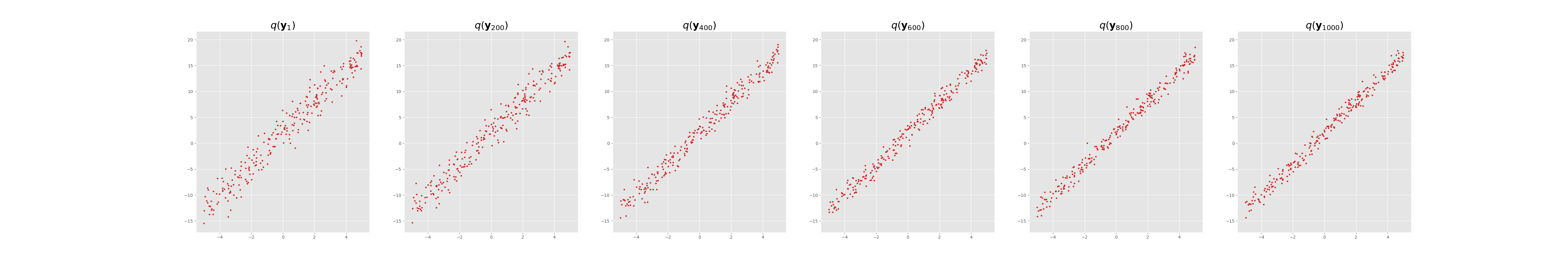}
    }
    \makebox[\textwidth][c]{
      \includegraphics[width=1.2\textwidth]{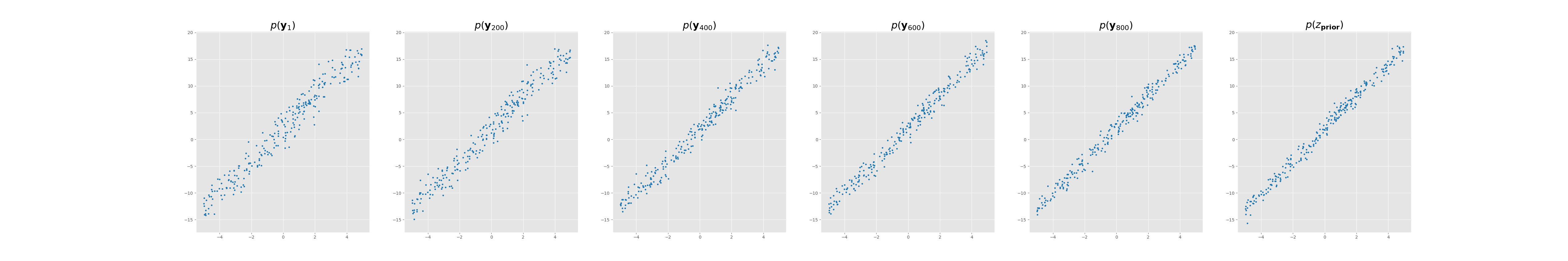}
    }
    \vspace{-6mm}
    \caption{$q$ and $p$ distribution samples for linear regression task during training. (\textbf{Top}) left to right: $q\big(\vy_t\given\vy_0, f_{\phi}(\vx)\big)$ for $t=1,200, \dots, T$; (\textbf{Bottom}) right to left: $p_{\theta}(\vy_{t-1}\given \vy_t, \vx)$ for $t=T, \dots, 200, 1$.}
    \label{fig:q_p_samples_linear_reg}
\end{figure*}

We present the evolution of both $q$ and $p$ distribution samples through the forward and reverse diffusion process, respectively. We first visualize the behaviors of these samples from the training on linear regression tasks in Figure \ref{fig:q_p_samples_linear_reg}, where we pick timesteps with an interval of $200$ steps including the $1^{st}$ and the last timestep, namely $t=1,200, 400, 600, 800, T$. The $p$ samples presented are from near the end of training. We observe that $\epsilonv_{\rvtheta}$ has been trained to match the $q$ samples at different timesteps well, including the variance. Furthermore, note that the true variance from the data generating function is set to $4$, while the prior $p(\vy_T\given\vx)$ has a variance of $1$. We can observe the gradual increase of variance in the reverse direction. This example helps to illustrate that when $f_{\phi}(\vx)$ can already estimate the mean accurately, it makes the task for the diffusion model easier: in this case to solely focus on recovering the aleatoric uncertainty.

\begin{figure*}[ht]
    \vspace{-3mm}
    \makebox[\textwidth][c]{
      \includegraphics[width=1.2\textwidth]{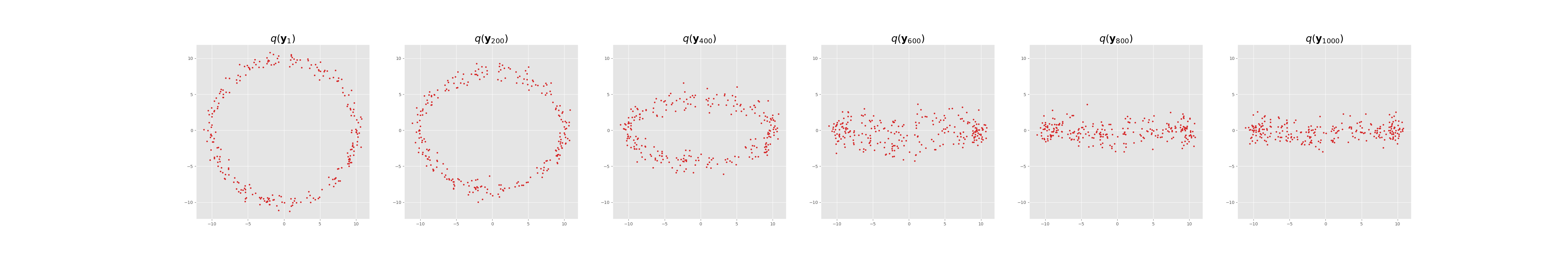}
    }
    \makebox[\textwidth][c]{
      \includegraphics[width=1.2\textwidth]{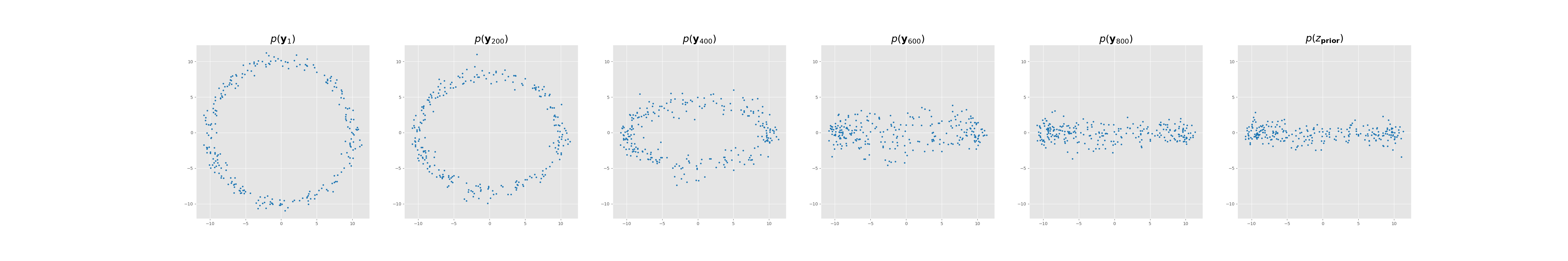}
    }
    \vspace{-6mm}
    \caption{$q$ (top) and $p$ (bottom) distribution samples for full circle regression task during training.}
    \label{fig:q_p_samples_full_circle}
\end{figure*}

\begin{figure*}[ht]
    \vspace{-3mm}
    \makebox[\textwidth][c]{
      \includegraphics[width=0.4\textwidth]{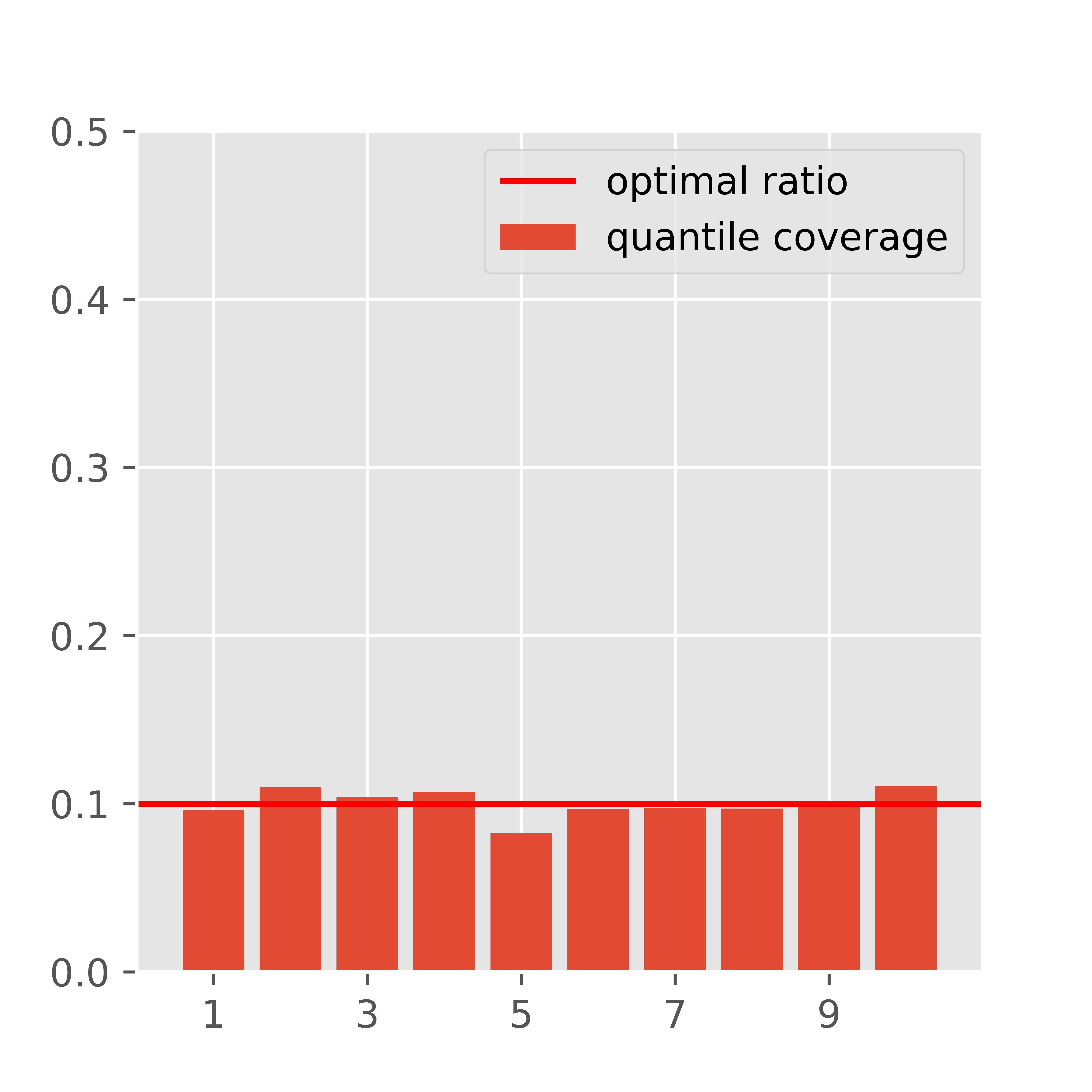}
    }
    \vspace{-6mm}
    \caption{Sample coverage ratio by bins for full circle regression task (QICE $0.62$).}
    \label{fig:qice_vis}
\end{figure*}

Similarly, we present the samples from $q$ and $p$ distribution during training for the full circle regression task in Figure \ref{fig:q_p_samples_full_circle}. Besides observing the matching in samples at all selected timesteps, we emphasize CARD's ability to model multi-modality at various intensities. As $t$ increases, we observe the samples gradually evolve from a uni-modal distribution into a multi-modal one, and the diffusion model is able to capture such progress. To quantify such match, we plot the quantile coverage ratios for samples at $t=0$ from one run, along with the optimal coverage ratio ($0.1$, for $10$ bins), in Figure \ref{fig:qice_vis}. Note that we obtain a QICE of $0.62$ in this run. We observe that CARD samples cover the true data with a ratio close to the optimal across all bins; The $5^{th}$ bin has relatively the most deviation from the optimal ratio (which is understandable as the full circle dataset has a bi-modal distribution across most $\vx$, with no data points in the center portion), which is compensated by the $2^{nd}$, $4^{th}$ and last bin with coverage slightly above the optimal ratio.

\begin{figure*}[ht]
    \vspace{-3mm}
    \makebox[\textwidth][c]{
      \includegraphics[width=1.2\textwidth]{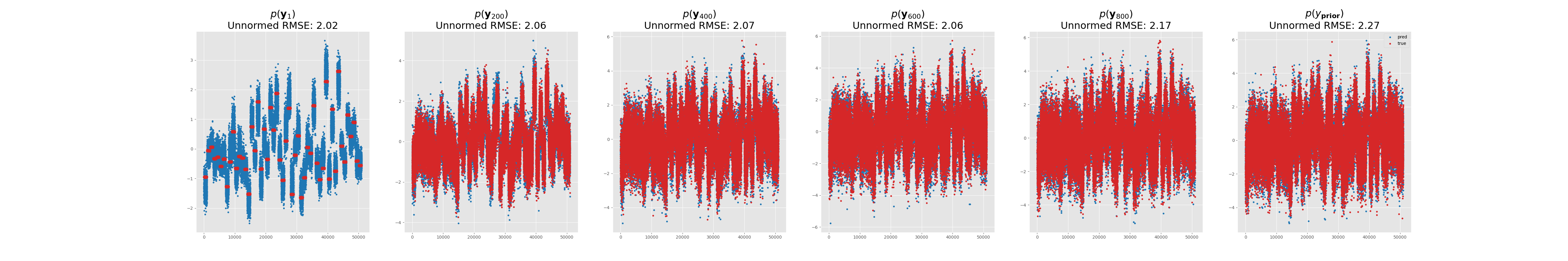}
    }
    \vspace{-6mm}
    \caption{$q$ (red) and $p$ (blue, $1000$ samples per $\vx$) samples from UCI Boston test set.}
    \label{fig:q_p_samples_boston_test_split_5}
\end{figure*}

We continue with a plot of samples during test time from the UCI Boston dataset. In Figure \ref{fig:q_p_samples_boston_test_split_5}, we plot the generated samples from $p$ (in blue) along with $q$ samples (in red) at various $t$. The $x$-axis represents the count of samples, instead of the actual $\vx$ since the true covariate space is high-dimensional. Note that we generate $1000$ samples given each $\vx$. While still observing the good mix between $q$ and $p$ samples for $t=200,\dots,T$ from $2^{nd}$ plot to the right, we observe that for $t=1$, all samples for each $\vx$ forms a vertical region that covers the corresponding $\vy_1$ (which shall be very close to $\vy_0$ due to the linear $\beta_t$ schedule) from $q$ distribution at various positions (\textit{i.e.}, middle, upper half, lower half, near top, near bottom). We observe that the samples are representative of the true conditionals $p(\vy\given\vx=x)$ for each $x$.

\subsection{The Development of Model Performance through the Reverse Diffusion Process}
\begin{figure*}[ht]
    \vspace{-3mm}
    \makebox[\textwidth][c]{
      \includegraphics[width=\textwidth]{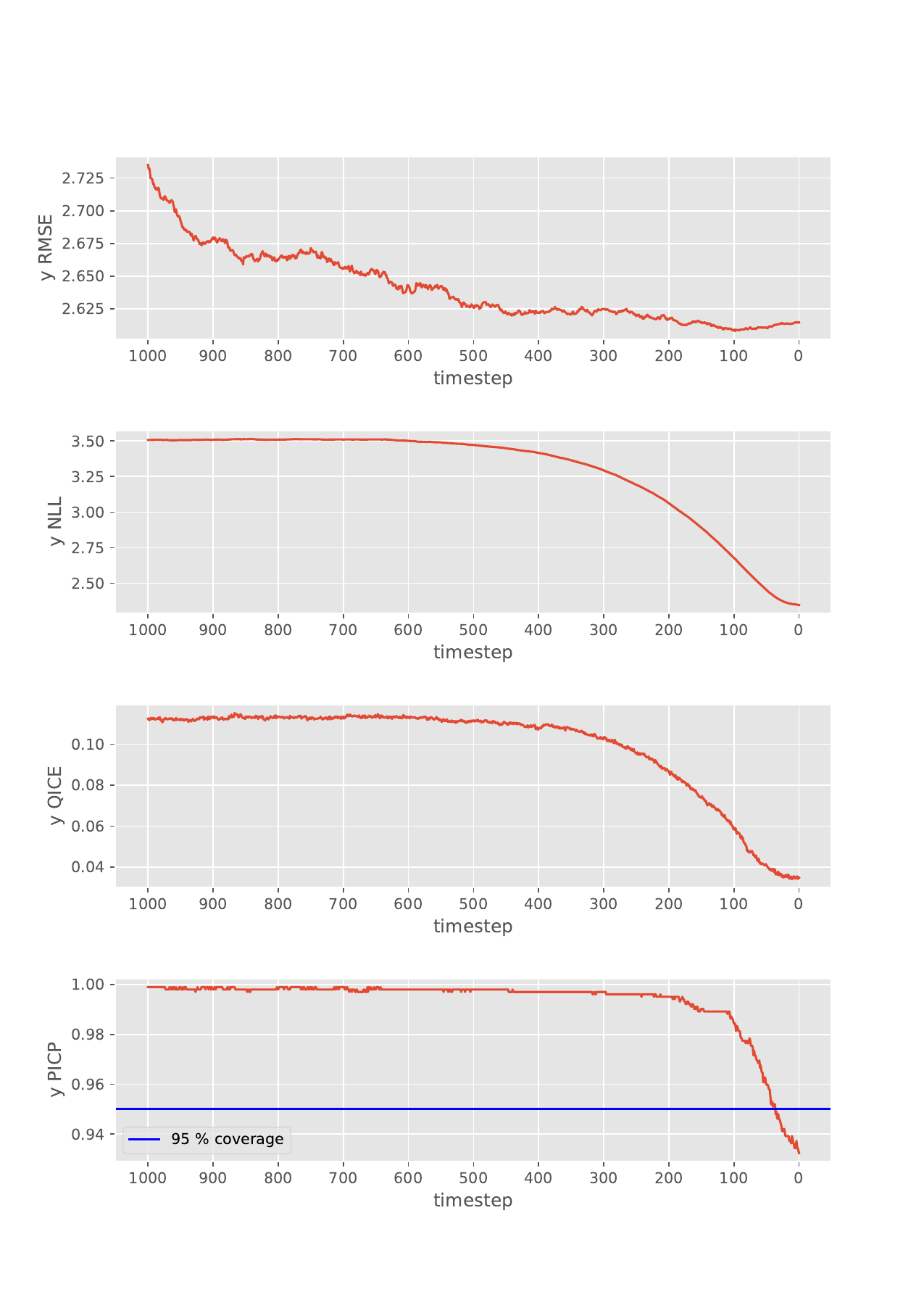}
    }
    \vspace{-6mm}
    \caption{Change in regression evaluation metrics on UCI Boston dataset. The value at each timestep is the mean across all $20$ splits.}
    \label{fig:metrics_all_splits_all_t_boston}
\end{figure*}

Following the demonstration of behavior change in samples with respect to timestep $t$ during both the training and test time, we present the change in model evaluation metrics on the test set as a function of $t$. We again use the UCI Boston dataset, and compute all evaluation metrics, including RMSE, NLL, QICE, and PICP, at all timesteps from $t=T$ to $t=0$. Since the standard procedure for running the task contains $20$ different splits on the dataset, we compute these metrics at all timesteps for all splits, and take the mean across all splits at each timestep $t$. We combine these plots in Figure \ref{fig:metrics_all_splits_all_t_boston}. For RMSE, we observe that the performance at $t=T$ is already quite good, due to the setting of $f_{\phi}(\vx)$ as the prior mean; as the reverse process continues, the metric gradually improves by decreasing. Both NLL and QICE behave in a similar fashion: as $t$ decreases, the metric steadily improves. For PICP, note that it crosses the optimal value of $0.95$ coverage ratio around the end of the reverse process; however, it did not deviate much from such value. This plot demonstrates the successive improvements across all metrics during the reverse diffusion process, starting from an already decent place (in terms of RMSE and NLL) due to the application of an informative prior.

\subsection{Improving the Granularity of ECE from Subgroup to Instance Level}\label{ssec:ece_discussion}
In this section, we first present the definition of ECE. This material is from \citet{calibration}, and we include it here for completeness. We then provide our analysis, specifically about its granularity to measure prediction confidence by the model, which motivates us to introduce an alternative way to measure model prediction confidence at a finer granularity (\textit{i.e.}, at instance level) in our paper.

ECE is defined as:
\ba{
    \text{ECE}\coloneqq\mathbb{E}_{\hat{P}}\big[\big|\mathbb{P}\big(\hat{Y}=Y\given\hat{P}=p\big)-p\big|\big], \label{eq:ECE}
}
where $Y$ and $\hat{Y}$ are true and predicted class labels, respectively; $\hat{P}$ is the predicted probability associated with $\hat{Y}$. A perfect calibration is defined as:
\ba{
    \mathbb{P}\big(\hat{Y}=Y\given\hat{P}=p\big)=p, \text{ }\forall p\in[0,1]. \label{eq:perfect_calibration}
}
However, since the predicted probability $\hat{Y}$ is continuous in $[0,1]$, we cannot compute ECE with finite instances, thus we approximate it by first \textbf{dividing the probability space into $M$ bins with equal width}, then compute the confidence and accuracy within each bin. Each test instance is placed into one specific bin by the predicted probability value associated with the true class label. For the $m$-th bin $B_m$, we have accuracy (proportion of correct predictions)
\ba{
    \text{acc}(B_m)=\frac{1}{|B_m|}\sum_{i\in B_m}\mathbb{1}(\hat{y}_i=y_i) \label{eq:bin_acc}
}
and confidence (mean of predicted probabilities)
\ba{
    \text{conf}(B_m)=\frac{1}{|B_m|}\sum_{i\in B_m}\hat{p}_i, \label{eq:bin_conf}
}
where $\hat{y}_i$ and $\hat{p}_i$ are the predicted label and its associated probability value, and $y_i$ is the true label, for instance $i$. We thus have the empirical version of ECE as:
\ba{
    \text{ECE}\coloneqq\sum_{m=1}^M\frac{|B_m|}{n}\big|\text{acc}(B_m)-\text{conf}(B_m)\big|, \label{eq:empirical_ECE}
}
where $|B_m|$ and $n$ are the cardinality of the $m$-th bin and the total number of instances, respectively.

To summarize, although we are interested in measuring the miscalibration through the difference between $p(y_i = \hat{y}_i\given\vx_i)$ and $\hat{p}_i$, we are only able to compute such miscalibration \textit{at the granularity of subgroup level} --- usually with the number of subgroups $M$ set to $10$ or less in practice. In other words, we cannot make a statement with the existing classification framework like the following: given this new test instance, we predict the class label to be (some class), but we are very sure our prediction is correct. This observation motivates us to introduce an alternative way of measuring model confidence, \textit{at the granularity of the instance level}.

\subsection{Insights on How CARD Accurately Recovers the Conditional Distributions}\label{ssec:card_insights}
Although a theoretical justification is not within the scope of our paper, in this section we plan to briefly talk about what makes CARD stands out as a great candidate to model $p(\vy\given\vx,\mathcal{D})$, through the capability to model implicit distributions by diffusion models in general: the theory of stochastic processes by \citet{feller1949theory} suggests that: 

a) With a large enough number of timesteps $T$, $q(\vx_T\given\vx_0)$ would converge to a stationary distribution $p(\vx_T)$ regardless of the distribution at timestep $t=0$, $q(\vx_0)$. In other words, we are able to go from any distribution $q(\vx_0)$ to a stationary distribution $p(\vx_T)$ by choice.

b) Meanwhile, with a large enough $T$ and small enough noise schedule $\{\beta_t\}_{t=1:T}$, the product in $q(\vx_{t-1}\given\vx_t)\propto q(\vx_t\given\vx_{t-1})q(\vx_{t-1})$ would be dominated by $q(\vx_t\given\vx_{t-1})$, thus both forward and reverse diffusion processes would share the same functional form. Although $q(\vx_{t-1}\given\vx_t)$ cannot be easily estimated, if we are able to learn a function $p_{\theta}(\vx_{t-1}\given\vx_t)$ that approximates $q(\vx_{t-1}\given\vx_t)$ well, we are able to reverse the direction mentioned in a) and go from $p(\vx_T)$ to any $q(\vx_0)$.

For real world datasets like the ones in \citet{ucidataset}, the relationship between the covariates and response variable can be quite complicated. The class of diffusion models in general, including CARD, places no restriction on the parametric form for $q(\vx_0)$ (\textit{i.e.}, it fits the data as it is), therefore it is suitable to the situations where a flexible distribution assumption is needed, instead of just the ones where an explicit distribution assumption is valid ($e.g.$, Gaussian for BNNs).

\subsection{Insights on Why CARD Outperforms Other BNN Approaches}
We observe the following two limitations of the class of BNN approaches: 

a) It places an explicit distributional form assumption to $p(\vy\given\vx,\bm{W})$, where $\bm{W}$ denotes the model parameter: for regression, it assumes an additive Gaussian noise model for $p(\vy\given\vx, \bm{W})$ (Eq. 2 in \citet{alex2017uncertainty}) --- we have addressed the limitation of such additive-noise assumption in the first two paragraphs of Section \ref{sec:intro}.

b) BNNs do not directly fit the actual posterior $p(\bm{W}\given\bm{X},\bm{Y})$, which is required for evaluating the marginal $p(\bm{Y}\given\bm{X})$, due to its intractability, but rather fit an approximated distribution $q(\bm{W})$ that minimizes its KL divergence to the actual posterior. Such distribution $q(\bm{W})$ has a simple distributional assumption (usually Gaussian), which places another layer of restrictions for BNNs to model their predictive distribution $p(\vy\given\vx) = \int p(\vy\given\vx, \bm{W})p(\bm{W}\given\bm{X},\bm{Y}) d\bm{W}$. 
CARD is able to dodge both limitations through the modeling of implicit distributions, as elaborated in Section \ref{ssec:card_insights}.

\clearpage
\section{Broader Impact and Limitations}
We believe that our model has a practical impact in various industrial settings, where supervised learning methods have been increasingly applied to facilitate the decision-making process. For regression tasks, \textit{e.g.}, understanding the relationship between drug dosage and biographical features of a patient in the medical domain, and evaluating player performance given various on-court measurements, where the actual distribution of the real-world data is complicated and unknown, we are able to compute summary statistics with a minimal amount of assumptions, as we do not assume the parametric form of the conditionals $p(\vy\given\vx)$. The analytical results from our framework thus have the potential to reach a broader audience, who in the past might be hindered by the sheer amount of jargon due to the lack of statistical training. For classification tasks, as mentioned in both of the experiments, we could easily adopt a human-machine collaboration framework: since our model is capable of conveying the prediction confidence along with the prediction itself, we could pass the cases where the model is less assertive to humans for further evaluation. This trait is especially valuable for classification tasks with exceptionally imbalanced data, \textit{e.g.}, fraud detection, and ad click-through rate prediction, where the volume of one class could be orders of magnitude more than the other. False negative errors for these applications are usually quite expensive, and simply adjusting the classification threshold would often put too many positive predictions for human agent evaluation. As demonstrated in our experiments, CARD is capable of providing uncertain cases with a very reasonable ratio, keeping the workload for human agents at a sensible level.

Meanwhile, since CARD is capable of modeling multi-modality, we are concerned that it could be devised for malicious purposes, like revealing personalized information of the patient through reverse engineering the prediction results: as an extension to the medical example of breast cancer given in the introduction section, one could tell with high confidence the gender of the patient based on the predicted mode. For research purposes, in this work, we only consider Gaussian diffusion and the reverse denoising in CARD, while there could be much more options when given different data. For example, in classification, we can optimize the classification likelihood with cross-entropy instead of using simple MSE loss, and directly perform diffusion in discrete spaces, as in \citet{austin2021structured}. We only take a pre-trained model $f_\phi$ as a deterministic neural network, while there could be more possibilities like combining BNN methods with CARD. Moreover, the computation efficiency of CARD may also be further investigated if the dataset size becomes larger. We encourage researchers in our community to further study those potential safety concerns and approaches for improvements in order to develop more mature supervised learning tools.

\section{Computational Resources}
Our models are trained and evaluated with a single Nvidia GeForce RTX 3090 GPU. We use PyTorch~\citep{pytorch}~1.10.0 as the deep learning framework. CARD trains between $100$ and $200$ steps per second at the batch size specified in Table \ref{tab:uci_batch_size_settings} for regression tasks, and at $44$ steps per second at batch size $256$ for classification on the CIFAR-10 dataset. The sampling for a batch of $250$ test instances, each with $100$ samples, takes around $1.05$ seconds at most for the classification task on FashionMNIST. The computation time could vary if we apply different network architectures for regression and classification tasks, and the architecture details are provided in Appendix \ref{ssec:general_experiment_details}.

\begin{table}[ht]
\centering
\caption{Model size and computation complexity of CARD and deterministic neural networks (architecture same as $f_\phi$) on different datasets. Throughputs measures the number of samples calculated per second. For UCI tasks, we measure the parameter size and computation complexity on the subset with the biggest data dimension.}\vspace{1mm}
\label{tab:complexity}
\renewcommand{\arraystretch}{1.0}
\setlength{\tabcolsep}{1.0mm}{ 
\setlength\tabcolsep{2pt}
\resizebox{\textwidth}{!}{
\begin{tabular}{l|ll|ll|ll|ll|ll}
\toprule[1.5pt]
Task    & \multicolumn{4}{c|}{Regression}                    & \multicolumn{4}{c}{Classification}                                                        \\ \hline
Dataset & \multicolumn{2}{c|}{Toy} & \multicolumn{2}{c|}{UCI} & \multicolumn{2}{c|}{FMNIST} & \multicolumn{2}{c|}{CIFAR-10/100}  & \multicolumn{2}{c}{ImageNet-100/1K} \\ \hline
model   & \#Params   & Throughputs  & \#Params   & Throughputs  & \#Params    & Throughputs    & \#Params     & Throughputs    & \#Params      & Throuputs     \\ \hline
CARD   & 5.24e-2M      & 25334.72  & 6.55e-2M      & 244146.47   & 6.41e-2M      & 1033.24      & 16.52M       & 244.21     & 32.76M       & 137.13        \\
DNN     & 5.22e-2M    & 25378.50   & 6.51e-2M   & 250555.79  & 6.02e-2M    & 1057.95      & 11.17M       & 249.56     & 25.62M        & 147.56        \\
\bottomrule[1.5pt]
\end{tabular}}
}
\end{table}

We summarize the model parameter size and measure the throughputs, \textit{i.e.}, how many data samples can be proceeded per second, as computation complexity measure in Table~\ref{tab:complexity}. As our model consists of both a diffusion model $\epsilonv_{\rvtheta}$ in the response prediction and a pre-trained prior mean model $f_\phi$, we compute them respectively.

\clearpage
\end{document}